\documentclass[10pt,twocolumn,letterpaper]{article}
\usepackage[utf8]{inputenc}
\usepackage{graphicx}

\title{Fitting 3D Shapes from Partial and Noisy Point Clouds\\with Evolutionary Computing}
\author{\hspace{-1.5cm}Jean F. Liénard\\\begin{minipage}{1.1\textwidth}\vspace{2cm}\hspace{-2cm}\includegraphics[scale=0.85]{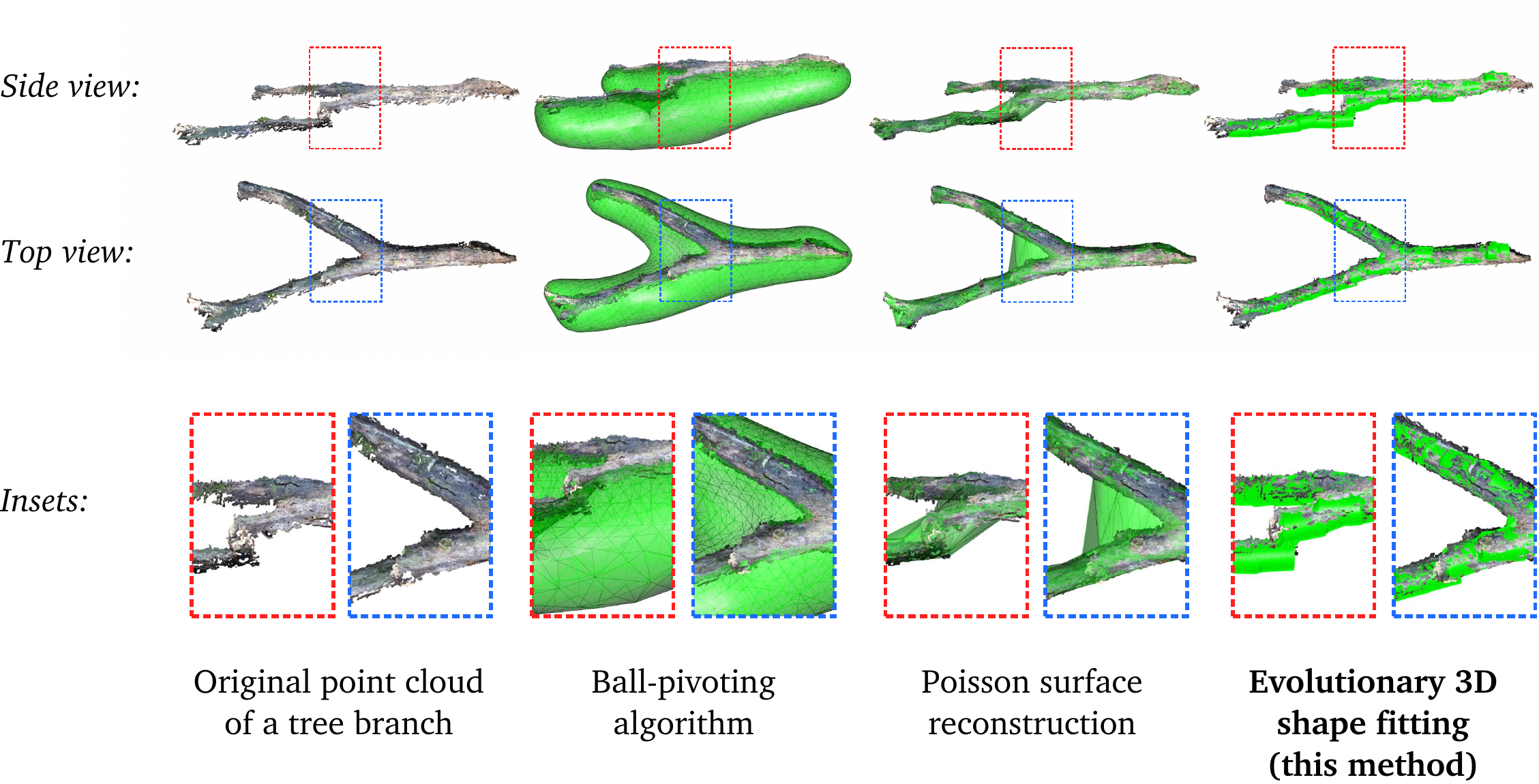}\vspace{1cm}\end{minipage}}

\usepackage{iccv}
\usepackage{times}
\usepackage{epsfig}
\usepackage{graphicx}
\usepackage{amsmath}
\usepackage{amssymb}

\iccvfinalcopy

\usepackage{algorithm2e}
\SetAlgorithmName{Pseudocode}{pseudocode}{List of Pseudocodes}

\usepackage[comma,authoryear]{natbib}

\usepackage{ctable}

\usepackage{floatrow}
\usepackage{subfig}
\usepackage{geometry}
\usepackage{mathtools}
\DeclarePairedDelimiter{\ceil}{\lceil}{\rceil}

\usepackage{setspace}

\begin{document}

\maketitle

\onehalfspacing

\begin{abstract}

Point clouds obtained from photogrammetry are noisy and incomplete models of reality. We propose an evolutionary optimization methodology that is able to approximate the underlying object geometry on such point clouds. This approach assumes \textit{a priori} knowledge on the 3D structure modeled and enables the identification of a collection of primitive shapes approximating the scene.
Built-in mechanisms that enforce high shape diversity and adaptive population size make this method suitable to modeling both simple and complex scenes.
We focus here on the case of cylinder approximations and we describe, test, and compare a set of mutation operators designed for optimal exploration of their search space. 
We assess the robustness and limitations of this algorithm through a series of synthetic examples, and we finally demonstrate its general applicability on two real-life cases in vegetation and industrial settings.
\onehalfspacing

\end{abstract}

\definecolor{lightgrey}{gray}{0.95} 

\fcolorbox{black}{lightgrey}{
  \begin{minipage}[t]{0.375\textwidth - 2\fboxsep}
\noindent\small Department of Mathematics and Statistics

\noindent\small Washington State University, Vancouver, USA

\noindent\texttt{jean.f.lienard@gmail.com}
  \end{minipage}
}

\normalsize

\section{Introduction}

Object reconstruction from three-dimensional point clouds, a process known as photogrammetry, has become widely available through theoretical breakthroughs and the release of software packages  \citep[e.g. Visual SFM][]{wu2011visualsfm}. In particular, 3D reconstruction has already found numerous applications in paleontology \citep{falkingham2012acquisition,falkingham2014historical}, and architecture and archaeology \citep{kersten2012potential}, or forestry \citep{Gatziolis_at_al_2015}.
As generating 3D point clouds from a collection of pictures becomes a streamlined process thanks to ready-to-use software \citep{wu2013towards}, the identification of geometric structures from discrete point clouds emerges as a challenging problem.

Historically, the oldest methods to tackle this problem were surface smoothness approach that rely on local parametric approximations of the point cloud, often assuming that it is free of noise \citep{berger2014state}. 
A wide array of general methods to reconstruct meshes from point clouds exist, including Ball pivoting \citep{bernardini1999ball}, Marching cubes \citep{lorensen1987marching}, Poisson surface reconstruction \citep{kazhdan2006poisson}, and the Alpha-hull approach \citep{edelsbrunner1983shape}.
These methods are being successfully applied in some domains, such as paleontology where the body volume of extinct species is estimated from convex hulls of the fossil bone structures \citep{sellers2012minimum}. 
However, generic-purpose mesh reconstruction techniques fail to deliver adequate approximations when parts of the scene are missing, or when the noise in a 3D point cloud is high. Their shortcomings are due to the fact that they make only minimal assumptions on the underlying shapes of the objects. 

In this work, we investigate another approach: enforcing geometrical assumptions about the scene to obtain suitable approximations of the mesh.
The core idea of this model-driven approach is to parameterize primitive shapes including spheres, cylinders, cones and toruses \citep{schnabel2007efficient}. This approach can be seen as the 3D analog to the 2D ``deformable template'' approach employed in object detection \citep{mesejo2016survey}.
Several reasons make this problem challenging from the optimization point of view. First, the search space has high dimensionality: an individual primitive shape, such as a cylinder, requires four or more parameters to encode its geometrical features\footnote{The simplest closed shape in space is a sphere and requires three parameters to encode the spatial coordinates of its center, and one parameter for its radius. More complex shapes, such as cylinders, require more parameters: two additional parameters for the orientation, and another one for the axial length.}, and each scene is composed of many primitive shapes. For example, the branch structure of a tree or a pipe-run network comprises dozens to hundreds of cylinders resulting in a set of solutions with more than one thousand parameters. Moreover, the 3D point clouds often presents with many artifacts, regardless of the reconstruction software used \citep{comparison_paper}: it is noisy and only visible aspects of the scene can be captured in the 3D point cloud. Furthermore, except for a few man-made objects, most primitive shapes will only approximate the real surface: for example, a tree branch is only cylindrical as a first approximation. These properties of 3D data make the search landscape filled with local optimization maxima (i.e. shapes with imperfect fits to the point cloud but for which any small -- or ``local'' -- modifications of their features, such as a small change in orientation or size, would result in worse fits). Finally, the cost function that evaluates the goodness-of-fit of a given shape to the scene has to be computed over the discrete set of 3D points, leading to a high computational burden every time a shape has to be assessed.

To date, most shape-fitting algorithms have focused on reducing the complexity of the search space by resampling 3D points into small clusters and performing a local optimization of a single shape on each cluster \citep{schnabel2007efficient}.
Limiting the number of points considered thus enables a very quick optimization, and successive iterations lead to a finer approximation. 
However, the faithfulness of the end result depends heavily on the clustering heuristics, and sub-optimal segmentation heuristics will result in sub-optimal recovering of the objects real structure. 
Typically, clustering heuristics are based on the similarity of point locations and their normals \citep{schnabel2007efficient}.
They perform usually well for man-made objects with regular geometries (for example where all cylinders have similar radii and well-separated axis orientations), but become imprecise when point clouds contain noise and/or are incomplete.
Global alignment procedures have been developed to overcome this limitation \citep[such as GLOBFIT,][]{li2011globfit}, yet they still assume some global regularity in the scene, and are not robust to high levels of noise \citep{qiu2014pipe,berger2014state}.

More closely related to the approach that we develop here, cylinder-specific clustering procedures have also been investigated in previous works.
In particular, it was observed that normals of points representing cylinders form circles when projected on a Gaussian sphere, a property that can be exploited to facilitate clustering \citep{liu2013cylinder,qiu2014pipe}. 
While elegant, this heuristic performs well only when applied to straight cylinders oriented in distinctly separate directions, such as an industrial piping system, and it requires an ad-hoc procedure to model joints \citep{qiu2014pipe}. 
Another type of algorithm utilizes an iterative approach where cylinders are fitted in succession \citep{pfeifer2004automatic}. It has been successfully applied to modeling standing trees with cylindrical or closely related shapes \citep{raumonen2013fast,markku2015analysis}, although as with most greedy optimization processes it is highly dependent on the starting condition and is thus prone to convergence to local maxima.

Here we propose an evolutionary algorithm which fits a collection of shapes on a 3D point cloud. At odds with other approaches, this algorithm seeks to optimize simultaneously a population of cylinders approximating the scene as a whole -- without resorting to iterative cluster resampling \citep{schnabel2007efficient} or one-at-a-time shape optimization procedure \citep{pfeifer2004automatic}. This is made possible by relying on the evolutionary optimization paradigm, where a population of best-fitting cylinders is considered at every step. This algorithm avoids artifacts due to early segmentation, and enables convergence even in highly noisy or partial object representations.
Technically, our evolutionary framework capitalizes on two desirable optimization properties: elitism (best solutions are kept across generations) and diversity (solutions span the entire search space).
We also designed a collection of mutation operators that can be used to generate interesting variations of 3D shapes and thus explore their search space efficiently. To validate their robustness and practical relevance for 3D reconstruction, we adopt a framework derived from the game theory through the use of Shapley values \citep{shapleyValue}. This enables us to quantitatively test the contributions of individual mutation operators to the overall reconstruction success.
The performances of the algorithm are evaluated in a series of synthetic test cases made to exemplify typical problems with 3D point clouds (namely noise, partial object occlusion, and object geometry imperfectly matching the primitive shape). Finally, we present real-life experiments demonstrating successful mesh reconstruction in the context of vegetation modeling and industrial pipe-run network.

\section{Methods}

\subsection{Algorithm overview}

The goal of our algorithm is to obtain a set of shapes with comprehensive coverage of the 3D scene. The framework of evolutionary computation is suitable for this goal, as it allows to optimize a set (or \textit{population}) of solutions without any constraint on the fitness function \citep{holland1975adaptation}. The optimization strategy that we developed follows the canonical outline of evolutionary algorithms:
\begin{itemize}
    \item random initialization of the starting population
    \item until the termination condition is fulfilled, do:
    \begin{enumerate}
        \item select a subset of the population
        \item generate offsprings by applying mutation and cross-over operators
        \item score the new population fitness
        \item replace the old population with a new one, according to the fitness of individuals
    \end{enumerate}
\end{itemize}

We customize this general scheme to the specific case of optimizing a population of shapes spanning the entire 3D scene, with a focus on adapting the steps 2 and 3 above. Although our method is compatible with any parameterized shape, we selected to work with cylinders. Thus the mutation operators designed in Section \ref{sec:mut_ope} are tailored to these shapes. We also develop in Section \ref{sec:fitness} a fitness function with a built-in diversity mechanism that promotes the population's convergence toward spatially segregated geometrical shapes.

\subsection{Fitness function}
\label{sec:fitness}

\begin{figure}
    \centering
    \includegraphics[width=0.99\textwidth]{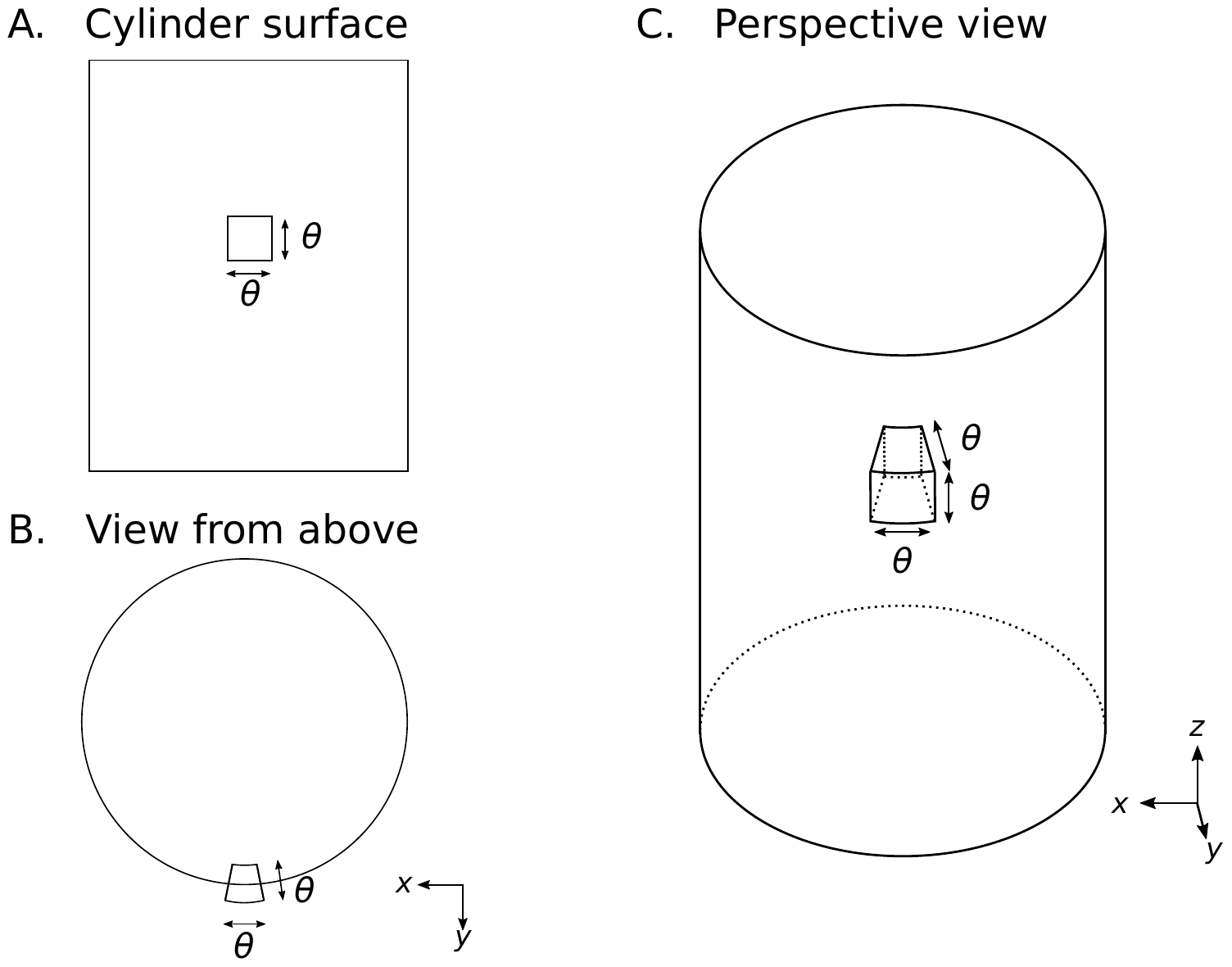}
    \caption{Illustration of the patch-based fitness. The potential fitness of a solution is computed as the proportion of patches extruded from the cylinder surface that contain points. Each patch is defined as a square of size $\tau\times\tau$ on the cylinder surface (panel A), which is then extruded along the radial axis by $\tau/2$ toward the inside and outside of the cylinder (panel B). Panel C shows one patch with an orthographic projection of the cylinder.}
    \label{fig:patch}
\end{figure}

The goal of the fitness function is to evaluate how closely each cylinder matches to the point cloud. When fitting a single shape, the mean distance to the point cloud is the metric of choice, coupled with a non-linear optimization framework \citep{lukacs1998faithful,shi2016genetic}. 
However this metric is not robust to missing data (i.e. where parts of the object are not represented in the point cloud). Also, it does not scale well to large and complex scenes where many points do not belong to cylinders.
Here we adopt another approach, in which we consider not the average distance to points of the scene, but the proportion of each primitive shape that is covered by points. To this end we discretize the primitive's surface at regular intervals into small patches of fixed size, and we count the fraction of patches having points in their immediate neighborhood (Fig. \ref{fig:patch}). With a patch size $\tau$, the number of patches along the circular axis of length $l$ is $i_\mathrm{max}=l/\tau$ (Fig. \ref{fig:patch}A), and along each circular cross-section this number is $j_\mathrm{max}=\frac{2\pi r}{\tau}$ (Fig. \ref{fig:patch}B). 
Formally, the binary function describing the occupancy of patch $(i,j)$ is given by:

\begin{equation}
\label{eq:ine_cond}
    \mathrm{filled} (i,j) = 1
    \Leftrightarrow
     \exists P / \left\{
    \begin{array}{l}
        \frac{2\pi r j}{j_\mathrm{max}} - \tau < \gamma < \frac{2\pi r j}{j_\mathrm{max}} + \tau\\
        r - \tau < y < r + \tau\\
        l \frac{i}{i_\mathrm{max}} - \tau < z < l \frac{i}{i_\mathrm{max}} + \tau
    \end{array}
    \right.,
\end{equation}

and the potential fitness $f$ of a cylinder is the proportion of filled patches:

\begin{equation}
\label{eq:filled}
f = \frac{\sum_{i=1}^{i_\mathrm{max}} \sum_{j=1}^{j_\mathrm{max}} \mathrm{filled}(i,j)}{i_\mathrm{max} . j_\mathrm{max}}
\end{equation}

The following terms: \textit{similar cylinders}, \textit{ideal cylinders} (relative to a point) and \textit{theoretical fitness} are employed to discuss our algorithm.
Many cylinders can have the same potential fitness function due to the discrete approximation of patches. 
We call these cylinders \textit{similar} as their fitness $f$ has the same value. Because similar cylinders are identical in terms of points overlap, we can arbitrarily pick one of them and discard all the others. 
We call the \textit{ideal} cylinder(s) for each point as the cylinder(s) with the highest potential fitness among all cylinders that include this point.
Finally, the \textit{theoretical fitness} $F$ is the potential fitness defined in Eq. \ref{eq:filled} but computed without the points already assigned to ideal cylinders that have a strictly higher theoretical fitness. In other words, the theoretical fitness of a cylinder is the fitness computed using only points that are not encompassed by another better-fitting cylinder. The theoretical fitness $F$ is thus equal to or lower than its potential fitness $f$. 

The theoretical fitness can not be used in the optimization procedure, as its computation relies on the knowledge of all the best-fitting cylinders - which is precisely the goal of the optimization procedure.
However, it is possible to compute an estimate of the theoretical fitness by considering only the set of ideal cylinders in the current population. We call this estimator the \textit{realized} fitness and denote it $\hat{F}$.
This quantity is used for solution ranking, leading to its maximization through the evolution's elitist selection. 
Conversely, as the population of shapes achieves increasingly good fits with the point cloud, the realized fitness becomes a better approximation of the theoretical fitness. 
This dual process results eventually in a collection of distinct shapes that cover the point cloud. 
The computation of the realized fitness is performed with the procedure described below:

\RestyleAlgo{boxruled}
\begin{algorithm}[ht]
 \Begin{
  \tcc{initialization}
  compute the patch occupancy for each point of each solution (Eq. \ref{eq:ine_cond})\;
  un-mark all solutions (a solution is marked when its realized fitness is computed)\;
 realized fitness list $\leftarrow \emptyset$\;
 \tcc{iterative computation of the realized fitness}
 \While{there are un-marked solutions}{
 \For{every un-marked solution $S_i$}{
  compute the potential fitness $f_i$, without the points already assigned to a marked solution (Eq. \ref{eq:filled})
  }
  identify $S_\mathrm{max}$ the solution with the highest potential fitness and mark it\;
  For $S_\mathrm{max}$, the potential fitness $f_\mathrm{max}$ is the realized fitness $\hat{F}_\mathrm{max}$: save it in the realized fitness list\;
 }
 }
 \caption{Computation of realized fitness
\label{alg}}
\end{algorithm}

This approach maintains diversity in a way that is conceptually similar to the clearing strategy \citep{petrowski1996clearing,petrowski1997new}, as only the best solutions are kept in each neighborhood. Its computation is efficient: while the establishment of the patch occupancy for a given cylinder is a computationally intensive task (it requires calculating the geometrical inequalities of Eq.~\ref{eq:ine_cond} after having expressed the points in the cylinder referential), it is independent of the other cylinders. Thus, the resource-demanding patch occupancy calculation needs to be performed only once for solutions kept across generations.

\subsection{Adaptive population size}
\label{sec:pop}

The number of shapes required to cover a point cloud is hard to estimate \textit{a priori}.
We can assume that any point of the scene will be in the neighborhood of a cylinder at some point of the optimization process, however in practice not all these cylinders should be retained (some points of the scene might be noise, or might belong to a non-cylindrical geometrical object). 
Hence we introduce the acceptance threshold $\alpha \in [0,1]$ which the user specifies as the minimal fractional coverage of each cylinder. This coverage depends on the density of the point cloud and on the object's representation completeness, and is investigated with synthetic examples (Fig. \ref{fig:noise} and \ref{fig:completeness}).

To ensure a complete exploration of the search space, we further enable dynamic growth and shrinkage of the population. This is achieved by indexing the population size on the number $n$ of solutions with coverage greater than $\alpha$. Prior to the offspring generation step, the new size of the population is computed as $max(\ceil{kn}, p)$, with $k>1$, and $p$ an arbitrary minimum population size. We established empirically that $p=50$ and $k=2$ are suitable settings, and we use these values in all the reconstructions presented in this paper.

\subsection{Mutation and Crossover operators}
\label{sec:mut_ope}

The choice of mutation operators suitable for exploring a 3D landscape depends heavily on the shape parameterization. Of the many different ways to parameterize shapes we chose a generic option applicable to most geometric primitives. First, we encode the shape position in space with the coordinates of its center ($x$, $y$, $z$). We then encode the shape direction using spherical coordinates consisting of two angles: the elevation $\theta \in [-\pi, \pi]$ and the azimuth $\phi \in [0, 2\pi]$. The shape length along its main axis (according to $\theta$ and $\phi$) is encoded by a positive number, $l$, and its radius is encoded by another positive number, $r$. These 7 parameters fully characterize a cylinder in the 3D space. Our approach can easily be extended to more complex shapes, such as cones, where the additional length parameters are treated similarly to $l$ and $r$ \citep{markku2015analysis}.

We designed four geometric transformation operators to enable spatial coherence during the exploration of 3D shapes:
\begin{itemize}
    \item \textbf{Translation}: mutate all spatial coordinates with
    $
    \left\{
    \begin{array}{l}
        x\leftarrow P_m(x)\\
        y\leftarrow P_m(y)\\
        z\leftarrow P_m(z)
    \end{array}
    \right.
    $

    \item \textbf{Rotation}: mutate the direction with
    $
    \left\{
    \begin{array}{l}
        \phi\leftarrow P_m(\phi)\\
        \theta\leftarrow P_m(\theta)
    \end{array}
    \right.
    $
        
    \item \textbf{Elongation}: mutate the length with $l\leftarrow P_m(l)$ 
    
    \item \textbf{Dilation}: mutate the radius with $r\leftarrow P_m(r)$
\end{itemize}
Where the bounded polynomial mutation operator $P_m$ introduced by \cite{deb1999niched} is used to update the value of the real-coded parameters. During the mutation step, each operator has a probability of $1/m$ to be selected and used, with $m$ the number of operators. Several operators can thus be combined in one mutation step to enable complex shape modifications.

\begin{figure}[t]
    \centering
    \begin{minipage}{0.32\textwidth}
        \sidesubfloat[position=bottom]{%
            \includegraphics[width=0.79\textwidth]{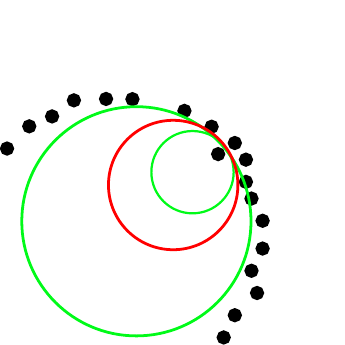}}
    \end{minipage}
    \begin{minipage}{0.32\textwidth}
        \sidesubfloat[position=bottom]{%
            \includegraphics[width=0.79\textwidth]{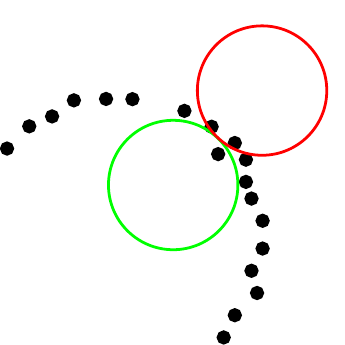}}
    \end{minipage}
    \begin{minipage}{0.32\textwidth}
        \sidesubfloat[position=bottom]{%
            \includegraphics[width=0.79\textwidth]{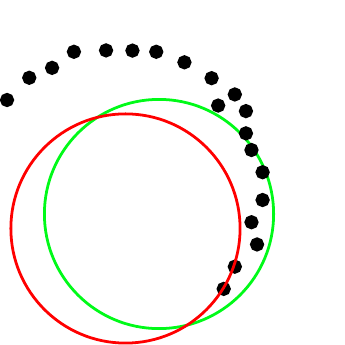}}
    \end{minipage}
    \caption{Illustration of the targeted mutation operators. The points to be approximated by primitive shapes are shown in black. The red circles correspond to the original position of cylinders, prior to mutation. The green circles illustrate possible outcomes of the three operators (A: Targeted Dilation, B: Targeted Flip, and C: Targeted Translation).}
    \label{fig:ope_mutation}
\end{figure}

While the above are in theory sufficient to fully explore the search space, in practice cylinders fitting could be stuck in local fitness optima. Concretely, these local optima are cylinders that overlap imperfectly with the point cloud, but for which small changes in orientation or size would result in a lower fitness score. 
We identified three typical local optima situations (illustrated in Fig. \ref{fig:ope_mutation}) and designed mutation operators to overcome them. These operators rely on the location of best contact between the cylinder and the point cloud, which has a low computational footprint given that the patch occupancy has already been computed for the fitness (Eq. \ref{eq:ine_cond}). Given $\{c_x, c_y, c_z\}$ the vector from the cylinder center to the point of best contact, the additional operators are:
\begin{itemize}
    \item \textbf{Targeted Dilation} (Fig. \ref{fig:ope_mutation}A) leaves the point of best contact intact but increases/decreases the radius of the cylinder:
    $
    \left\{
    \begin{array}{l}
        r'\leftarrow P_m(r)\\
        x\leftarrow x + (r-r') c_x\\
        y\leftarrow y + (r-r') c_y\\
        z\leftarrow z + (r-r') c_z\\
        r\leftarrow r'
    \end{array}
    \right.
    $
    
    \item \textbf{Targeted Flip} (Fig. \ref{fig:ope_mutation}B) performs a symmetrical translation with respect to the point of best contact:
    $
    \left\{
    \begin{array}{l}
        x\leftarrow x + 2 r c_x\\
        y\leftarrow y + 2 r c_y\\
        z\leftarrow z + 2 r c_z
    \end{array}
    \right.
    $
    
    \item \textbf{Targeted Translation} (Fig. \ref{fig:ope_mutation}C) translates the cylinder along its axis to match the point of best contact:
    $
    \left\{
    \begin{array}{l}
        x\leftarrow c_x \frac{r}{2} sin(\phi)*cos(-pi/2+\theta)\\
        y\leftarrow c_y \frac{r}{2} cos(\phi)*cos(-pi/2+\theta)\\
        z\leftarrow c_z \frac{r}{2} sin(-pi/2+\theta)
    \end{array}
    \right.
    $
    
\end{itemize}

We also adapted the crossover operators to 3D shape optimization. Genes for the crossover operation are selected using a multi-point design, where crossing points match the fundamental blocks of 3D phenotype \citep{de1992formal}. Four such fundamental blocks can be identified with our encoding of cylinder shapes: 1) the triplet of spatial coordinates {X,Y,Z}; 2) the pair of orientation vectors {$\phi$,$\theta$}; 3) the axial length $l$; and 4) the radius $r$. Within these blocks, updated values are obtained using the Simulated Binary Crossover operator \citep{deb1994simulated}.

\subsection{Quantifying the relevance of mutation operators}

Given the empirical nature of the design of the mutation operators, we sought to assess quantitatively their relevance to the overall optimization performance.
For this we used Shapley values, which are metrics originally developed in the field of game theory to score the contribution of each player (here, mutation operator) to coalitions \citep{shapleyValue,aumann1989game}. This is done by considering all possible teams of players and seeing how changing the team composition alters the outcome (score) of the game. Formally, given the fitness function $\hat{F}$ and the set of mutation operators $M$, the Shapley value $\zeta_i$ of the mutation operator $i$ is defined as:

$$
\zeta_i(v)=
\mkern-6mu\sum_{S \subseteq M \setminus \{i\}} \mkern-14mu\frac{|S|!\; (|M|-|S|-1)!}{|M|!}(f(S\cup\{i\})-f(S))
$$

\subsection{Real world 3D reconstructions}

We tested the algorithm on actual point clouds obtained from processing videos acquired around different targeted objects. We selected two different cases in which objects consisted of cylinders: (a) a vegetation reconstruction featuring coarse woody debris, where the recovery of vegetation dimensions is relevant for estimating biomass and volume \citep{Gatziolis_at_al_2015}, and (b) an industrial reconstruction, where the recovery of pipe-run is useful to mapping their network and possibly identifying space for further extensions \citep{qiu2014pipe,pang2015automatic}.
118 photos from the vegetation example were taken with the low-resolution, integrated camera of a smartphone and subsequently reconstructed with Visual SFM \citep{wu2011visualsfm}. The pipe-run example was imaged with a professional-grade camera in a video with 4K resolution at 30 Hz frame rate. 
254 high-quality frames were extracted from the video and processed with \cite{manual2014professional} using the ``Ultra High'' quality settings. 

\subsection{Code availability}

The code performing the evolutionary optimization and the analysis are written in R \citep{R}, with geometrical routines computing the fitness in C++ with Rcpp \citep{Rcpp,RcppArmadillo} and Armadillo \citep{sanderson2016armadillo,sanderson2018user} for speed. The full code is available at:
\footnotesize\texttt{https://github.com/jealie/3D\_cylinder\_evolution}\normalsize.

\section{Results}

\subsection{Mutation operators analysis}

\begin{figure*}[htpb]
    \centering
    \begin{minipage}{0.45\textwidth}
    \sidesubfloat[position=bottom]{%
        \begin{minipage}{0.7\textwidth}
            \includegraphics[width=0.99\textwidth]{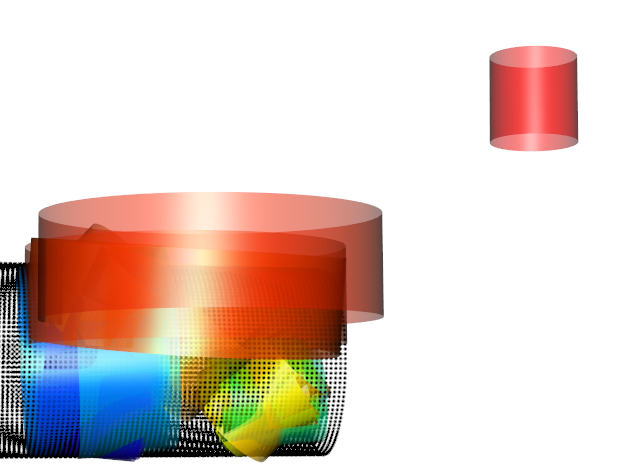}\\
    
            \includegraphics[width=0.99\textwidth]{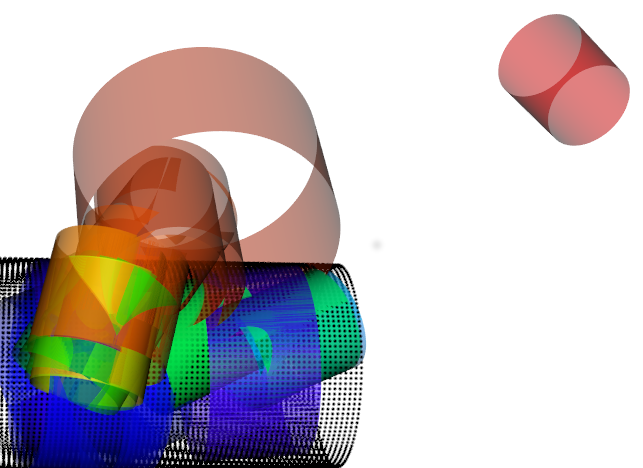}
        \end{minipage}
        \begin{minipage}{0.24\textwidth}
            \includegraphics[width=0.99\textwidth]{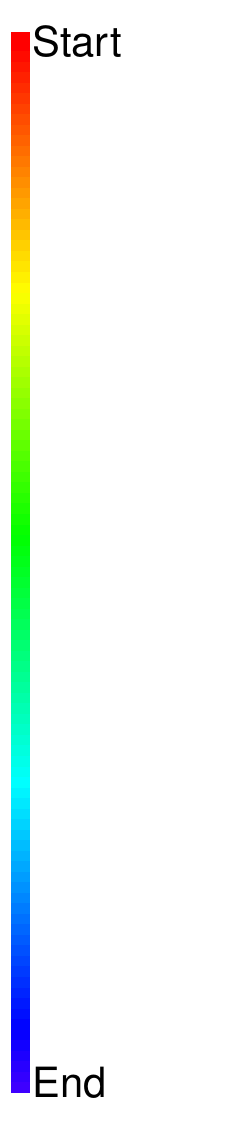}   
        \end{minipage}
        \hfill
    }
    \end{minipage}
    \begin{minipage}{0.5\textwidth}\sidesubfloat[position=bottom]{
        \includegraphics[width=0.99\textwidth]{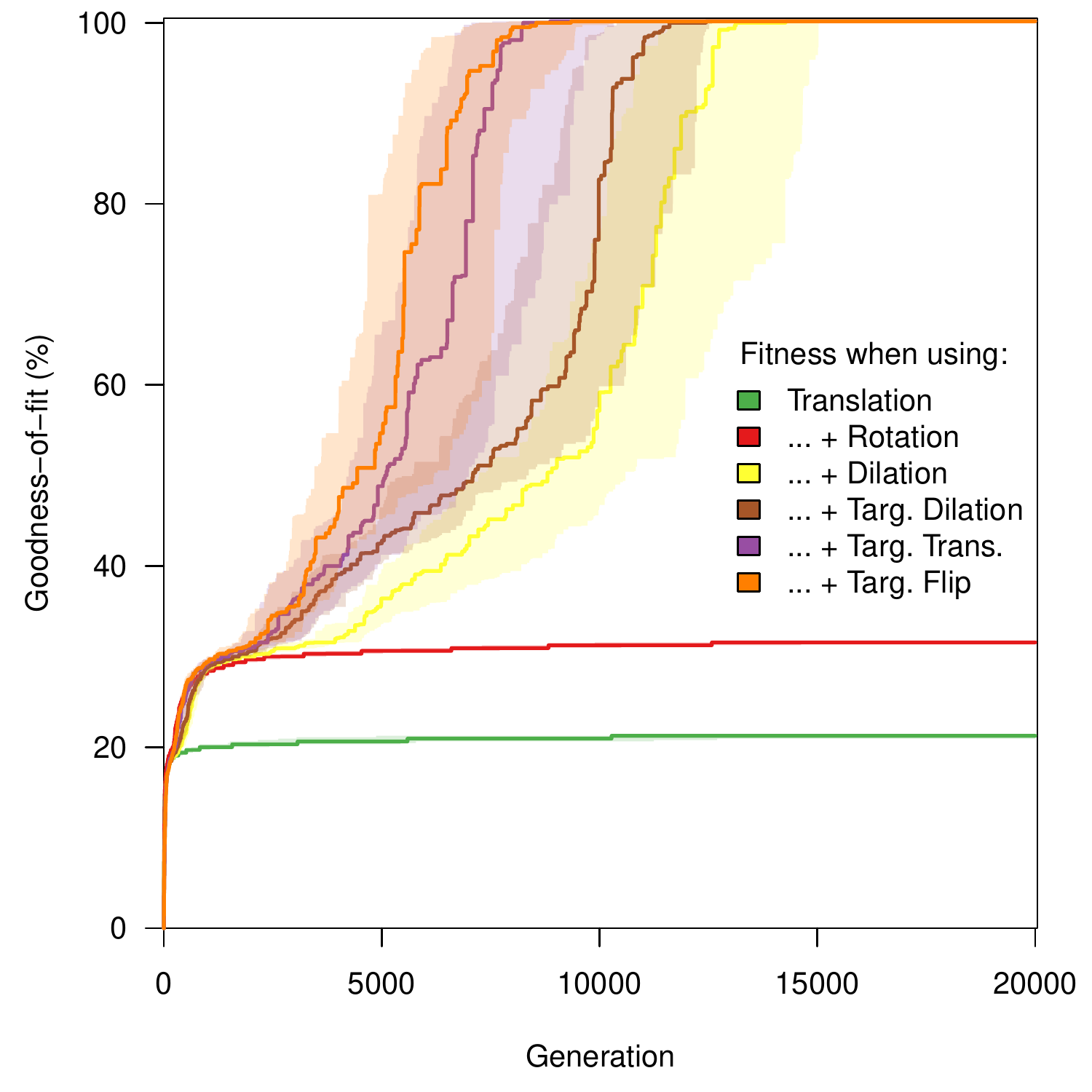}
        \hfill
    }\end{minipage}
    
    \caption{Illustration of the relative importance of mutation operators in the optimization process. In all the panels of this figure, the population size is limited to a singleton to focus on mutation operators (excluding the adaptive population size mechanisms). (A) Two examples show the evolution of the single solution when initialized in the upper-right corner of the point cloud. The chronological order of mutations is shown with colors. Black points indicate the target cylinder (the goal of the optimization). As the iterative process advances, which is reflected by a progression from red to blue hues, shapes with the proper orientation, placement and radius are found.
    (B) Goodness-of-fit when the mutation operators are successively enabled. The three operators with the highest Shapley value, Translation, Rotation and Dilation, suffice for a perfect fit (yellow trace). Enabling the Targeted mutation operators further decreases convergence time. Thick lines indicate the median performance, and shaded area show the bootstrapped 95\% confidence interval. 
    }\vspace{0.9cm}
    \label{fig:comp_ope}
\end{figure*}

We investigate the performance of 3D shape mutation operators using a simplified version of the evolutionary algorithm. In this version, the population is reduced to a singleton (i.e. a single solution), and thus the cross-over is omitted. We further designed a simplified task where the point cloud to approximate is regularly sampled on the surface of a single cylinder. Fig. \ref{fig:comp_ope}A shows two typical optimization runs, where the algorithm  explores the search space and eventually matches the target cylinder.

In this setting, the basic set of operators (translation, rotation, elongation and dilation) were sufficient to reach an optimal fit (Fig. \ref{fig:comp_ope}B). The extended set of targeted operators further sped up convergence, reducing the number of iterations by up to 50\% (Fig. \ref{fig:comp_ope}B).

The mutation operators explore the search space using different strategies, and their usefulness depends on the cylinder's position relatively to the point cloud. To investigate this spatial dependency in our analysis, we performed multiple optimization runs with different starting locations while enabling/disabling operators selectively (Fig. \ref{fig:comp_ope}C). The relative importance of mutation operators showed a strong spatial dependency. Not surprisingly, Translation and Rotation were the most important operators overall. Their relative importance changed as a function of the initial distance to the target cylinder, with Rotation being the most important operator when initialized inside the cylinder and Translation being the most important outside (Fig. \ref{fig:comp_ope}C). Dilation and Targeted Dilation were equally useful, especially when the search was initialized in the neighborhood of the cylinder. Targeted Translation and Targeted Flip were slightly less beneficial to the overall search quality (although enabling them led to quicker convergence, cf. Fig. \ref{fig:comp_ope}B).
The Elongation operator was not relevant in this task as the reference points formed a very long cylinder. The quantitative analysis revealed it to be neutral, or even detrimental, to the overall performance. This was expected in this particular setting where a singleton population is considered, and this operator remains essential as it is the only one that can change cylinder axial lengths.
Overall, this analysis demonstrates that the degrees of freedom granted by the chosen set of mutation operators is sufficient. It also demonstrates that the Targeted operators can improve convergence speed.

\subsection{Synthetic case studies}

\begin{figure*}
\vspace{2cm}
    \centering
    \begin{minipage}{0.39\textwidth}
    \sidesubfloat[position=bottom]{\includegraphics[width=0.49\textwidth]{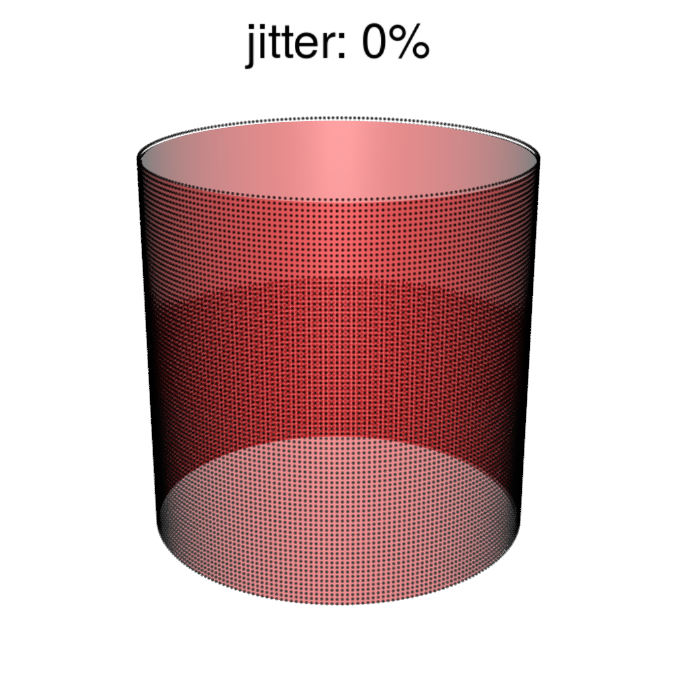}
    \includegraphics[width=0.49\textwidth]{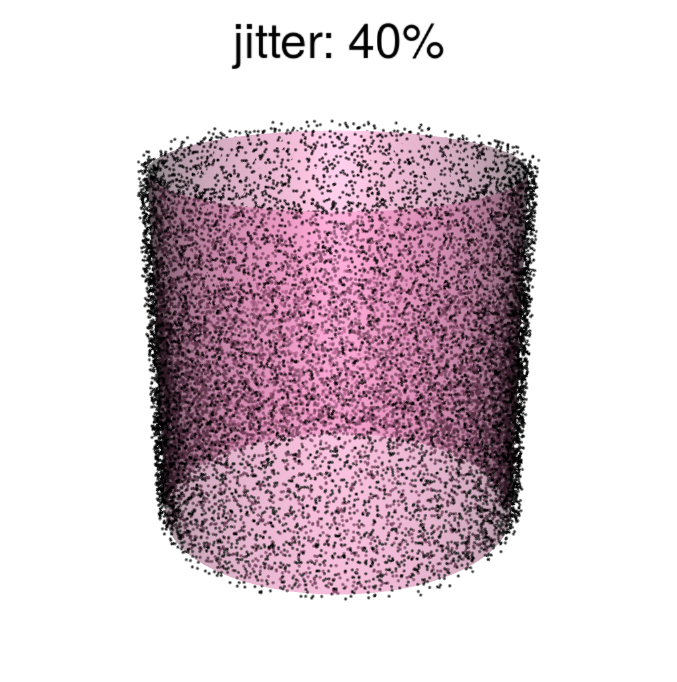}}
    
    \sidesubfloat[position=bottom]{\includegraphics[width=0.99\textwidth]{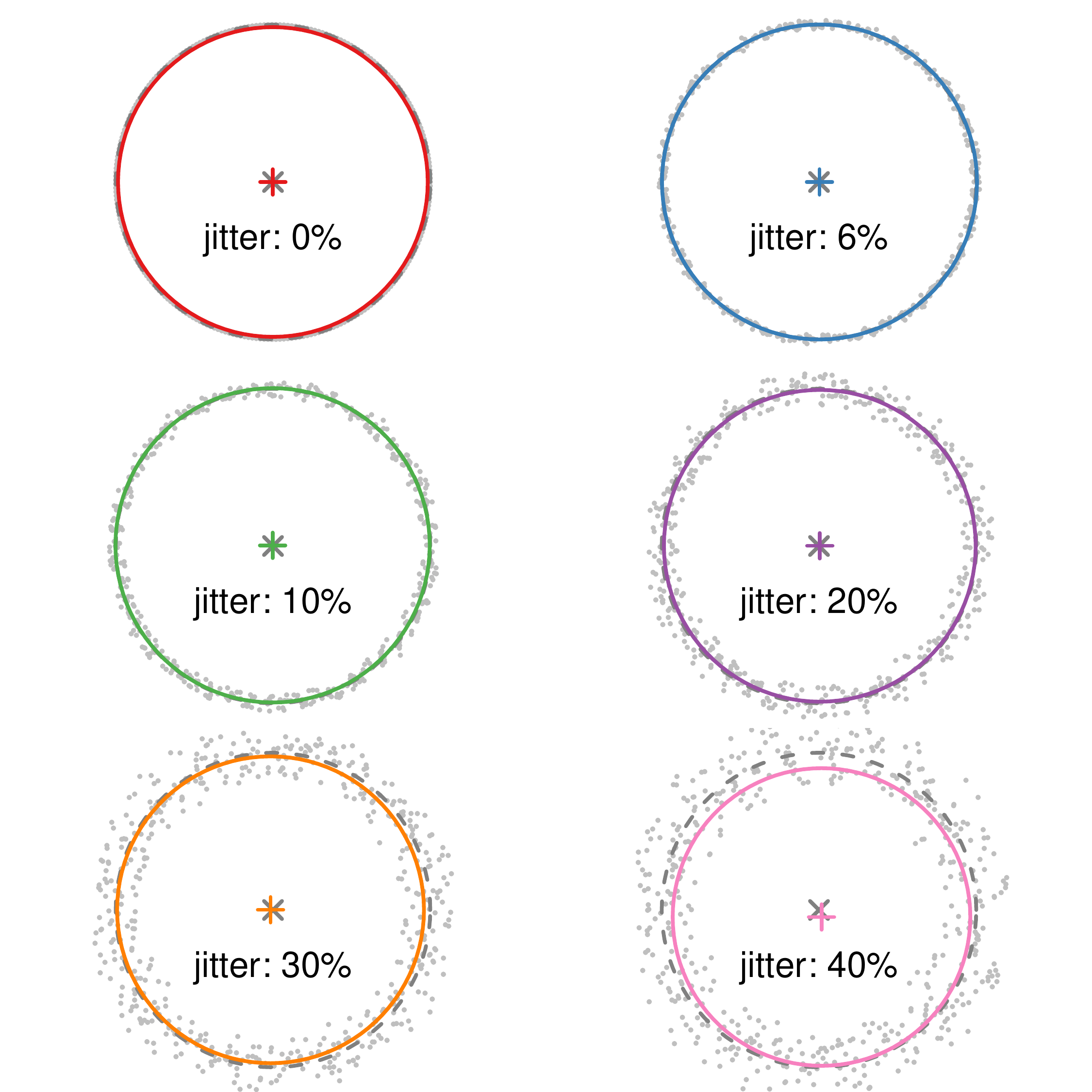}}
    \end{minipage}
        \begin{minipage}{0.59\textwidth}\sidesubfloat[position=bottom]{
\includegraphics[width=0.99\textwidth]{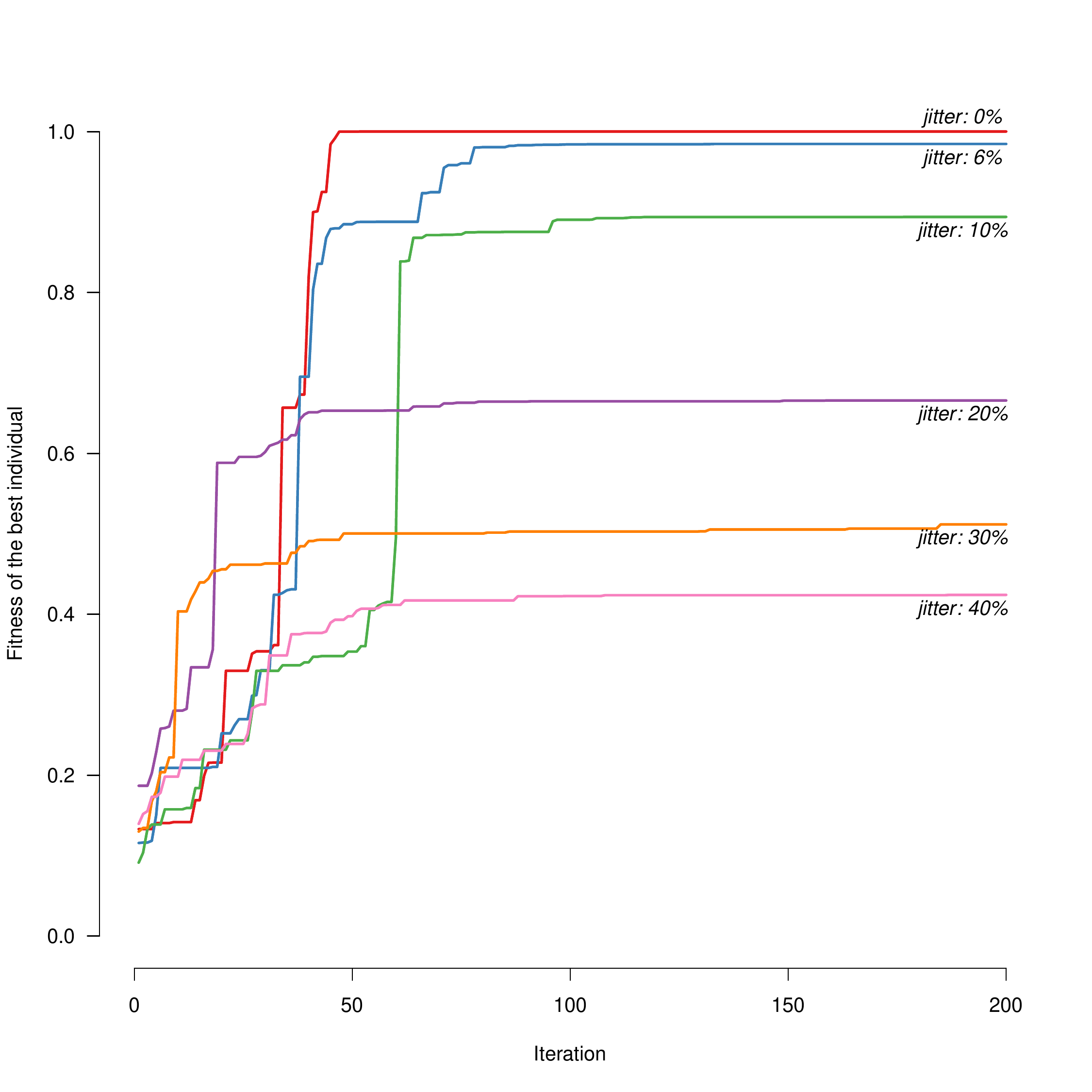}}
    \end{minipage}
    \caption{Performance of the cylinder search with added noise. (A) 3D views of the synthetic data and fitted cylinder when no (left) and substantial noise (right, 40\% jitter) is added. (B) Cross-sectional views with varying jittering intensity. Dashed circles and 'x' marks indicate the ideal solutions and their center. The colored circles and '+' marks represent the best fits. Near-perfect fits are achieved for noise level below 30\%. (C) Fitness scores of the best solution across the optimization process. The lower fitness values obtained with high jitter (panel C) reflect the poor overlap between optimal solutions and reference points the approximation is still nearly optimal as the excellent spatial correspondence in panel B indicate.}
    \label{fig:noise}
\end{figure*}

\begin{figure*}
\vspace{2cm}
    \centering
    \begin{minipage}{0.39\textwidth}
    \sidesubfloat[position=bottom]{\includegraphics[width=0.49\textwidth]{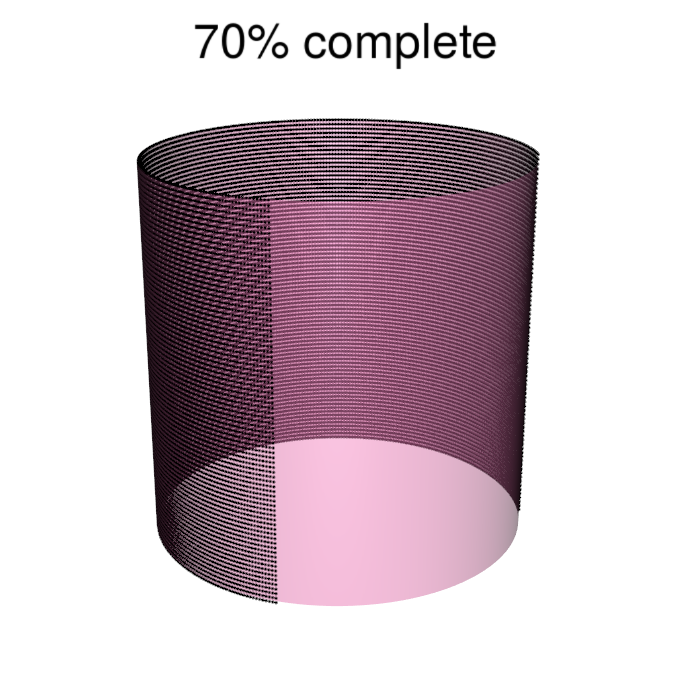}
    \includegraphics[width=0.49\textwidth]{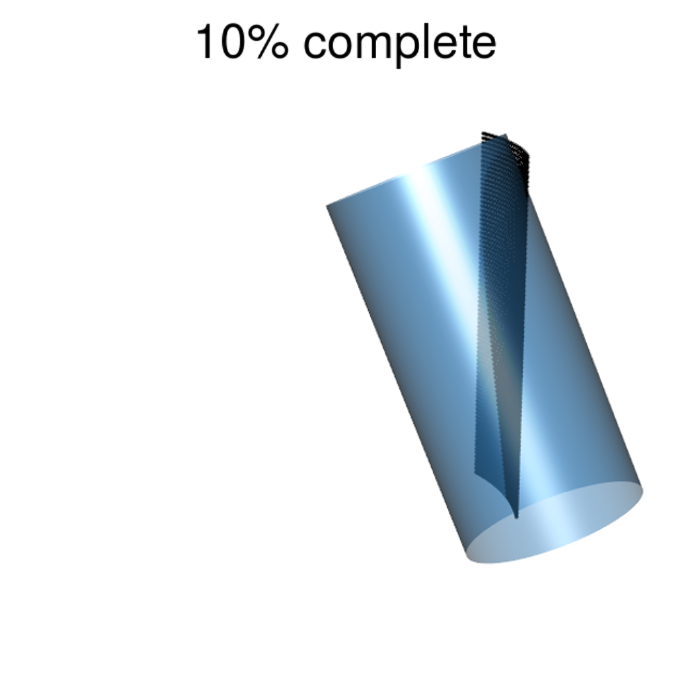}}
    
    \sidesubfloat[position=bottom]{\includegraphics[width=0.99\textwidth]{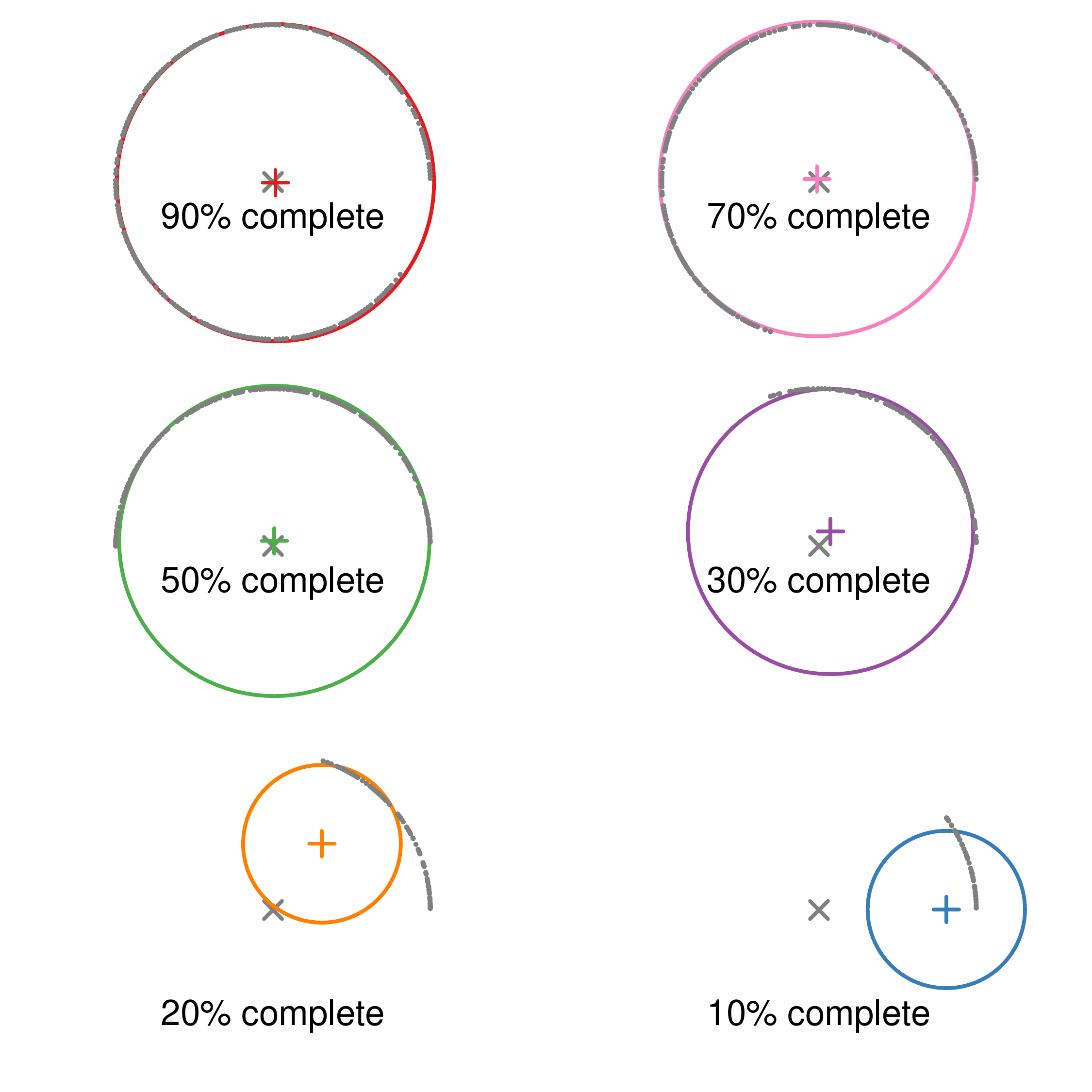}}
    \end{minipage}
        \begin{minipage}{0.59\textwidth}\sidesubfloat[position=bottom]{
\includegraphics[width=0.99\textwidth]{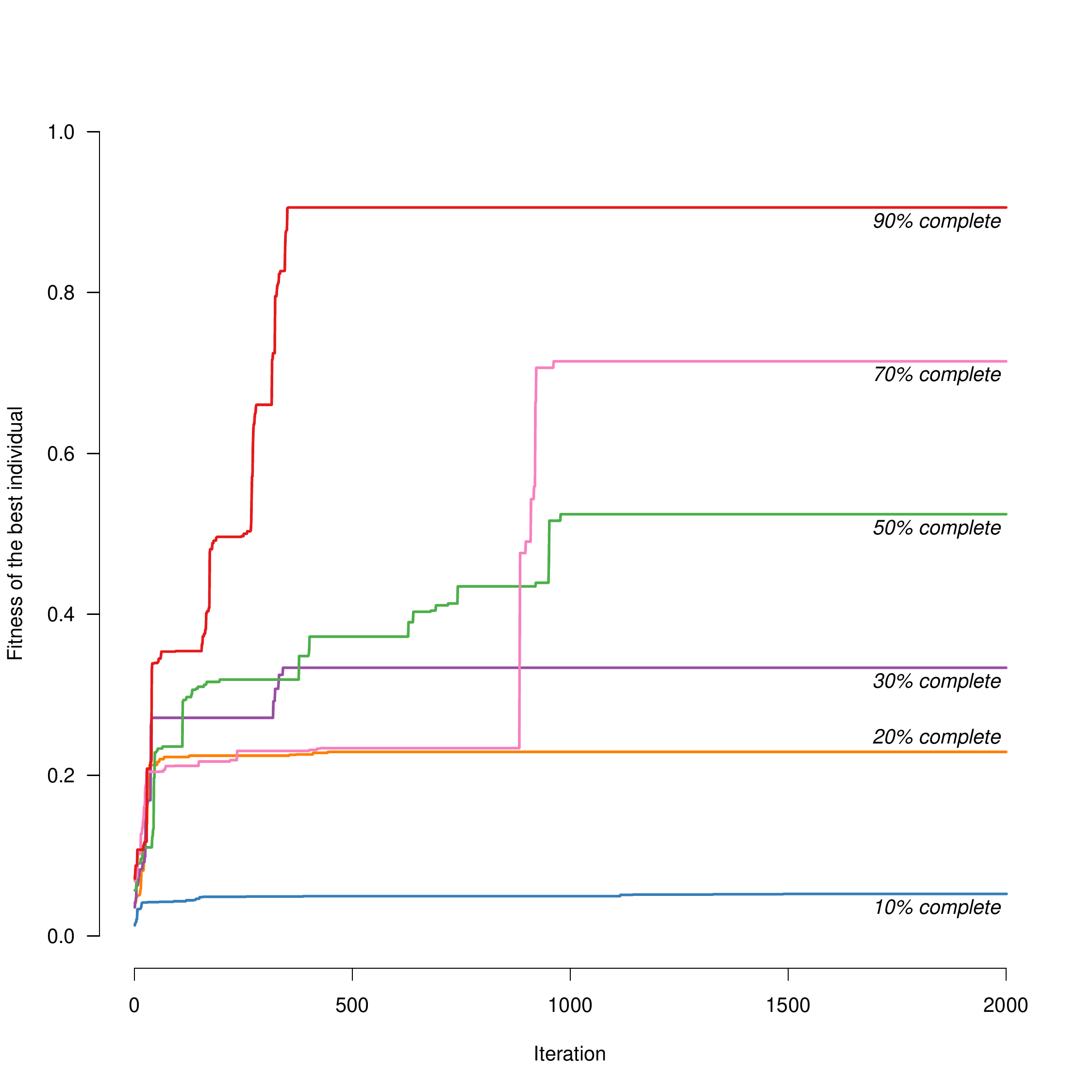}}
    \end{minipage}
    \caption{Performance of the cylinder search with partially represented cylinders. (A) 3D views of synthetic data and fitted cylinder with most (70\%) and a small fraction (10\%) of the surface present. (B) Cross-sectional views with completeness ranging from 10\% to 90\%. The optimized cylinders (colored circles) fail to recover the correct shape when the point cloud is too partial (10-20\% complete) but succeed when at least 30\% of the surface is represented. (C) Best solution fit across the optimization process. The fitness decreases with point cloud alterations, indicating poor overlap between the cylinder shape and the points (as in the added noise experiment of Fig. \ref{fig:noise}). However the spatial match remains high when 30\% or more of the cylinder's surface is represented in the original point cloud.}
    \label{fig:completeness}
\end{figure*}

\begin{figure*}
    \centering
    \sidesubfloat[position=bottom]{\includegraphics[width=0.49\textwidth]{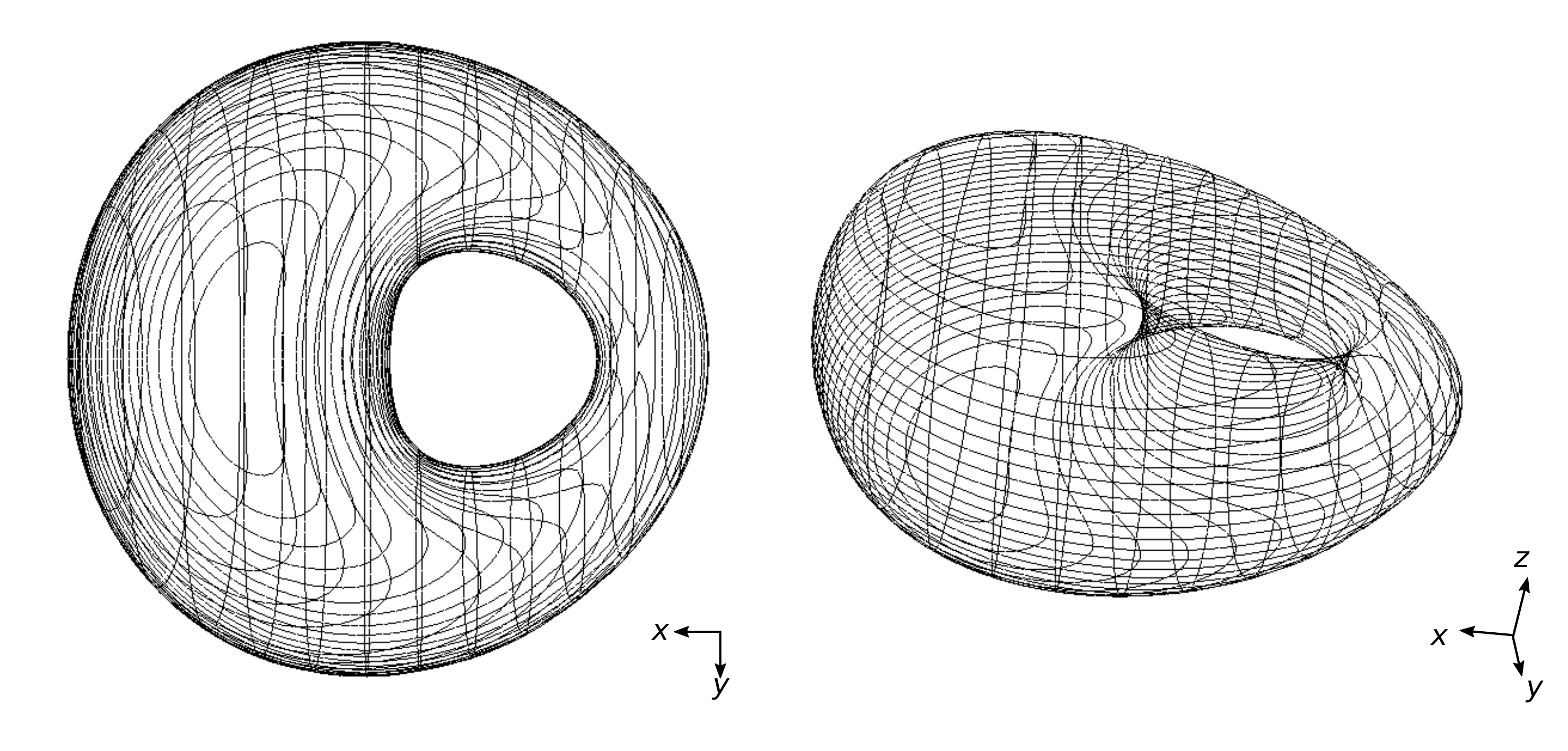}}
    \begin{minipage}[b]{0.49\textwidth}
    \sidesubfloat[position=bottom]{%
        \begin{minipage}{0.49\textwidth}%
                \includegraphics[width=0.99\textwidth]{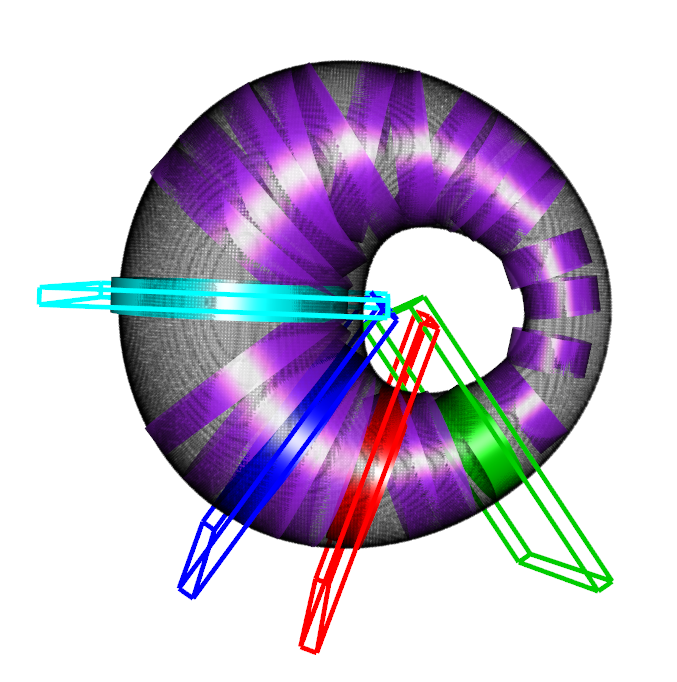}%
        \end{minipage}
        \begin{minipage}{0.49\textwidth}%
            \begin{minipage}{0.49\textwidth}%
                \includegraphics[width=0.99\textwidth]{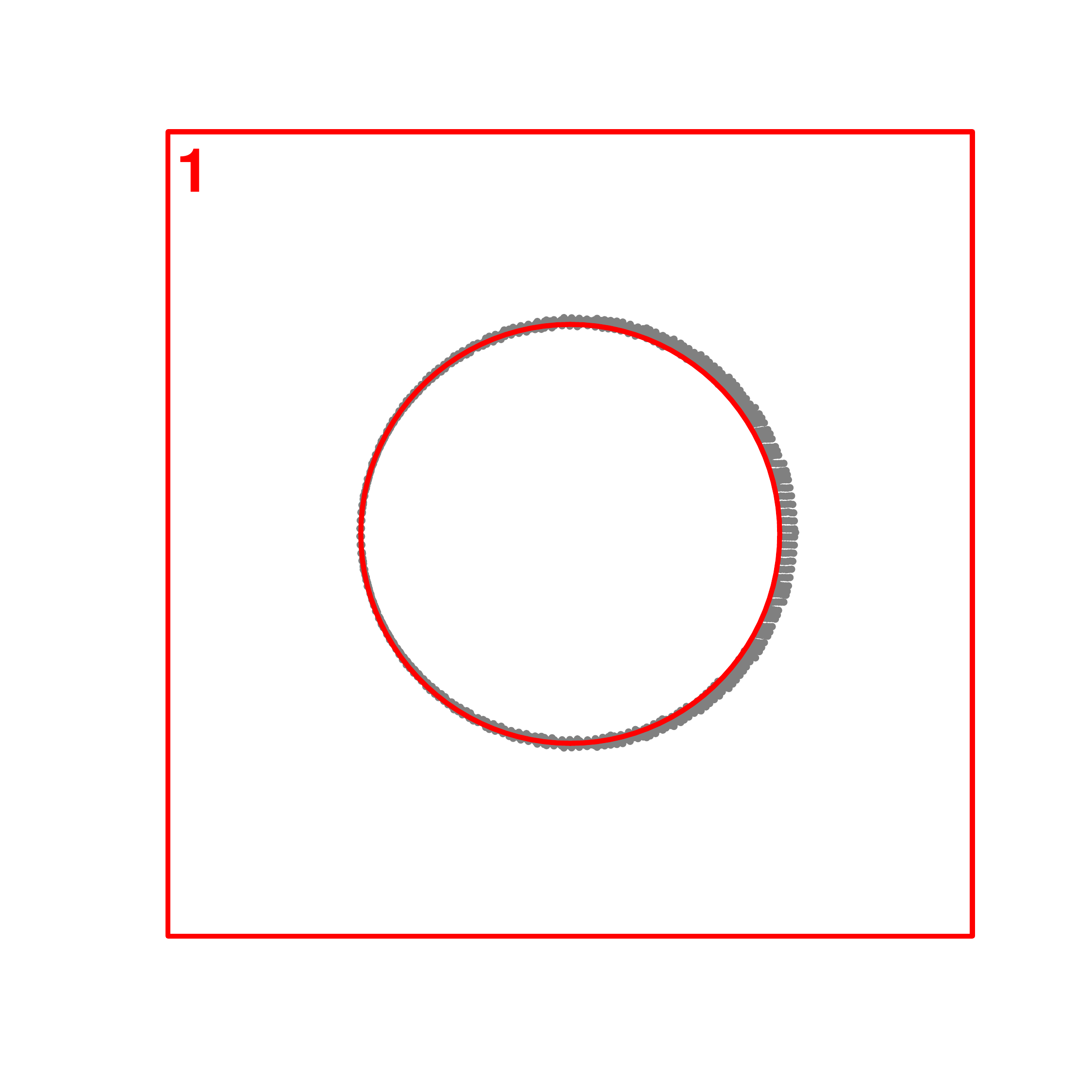}%
                \\%
                \includegraphics[width=0.99\textwidth]{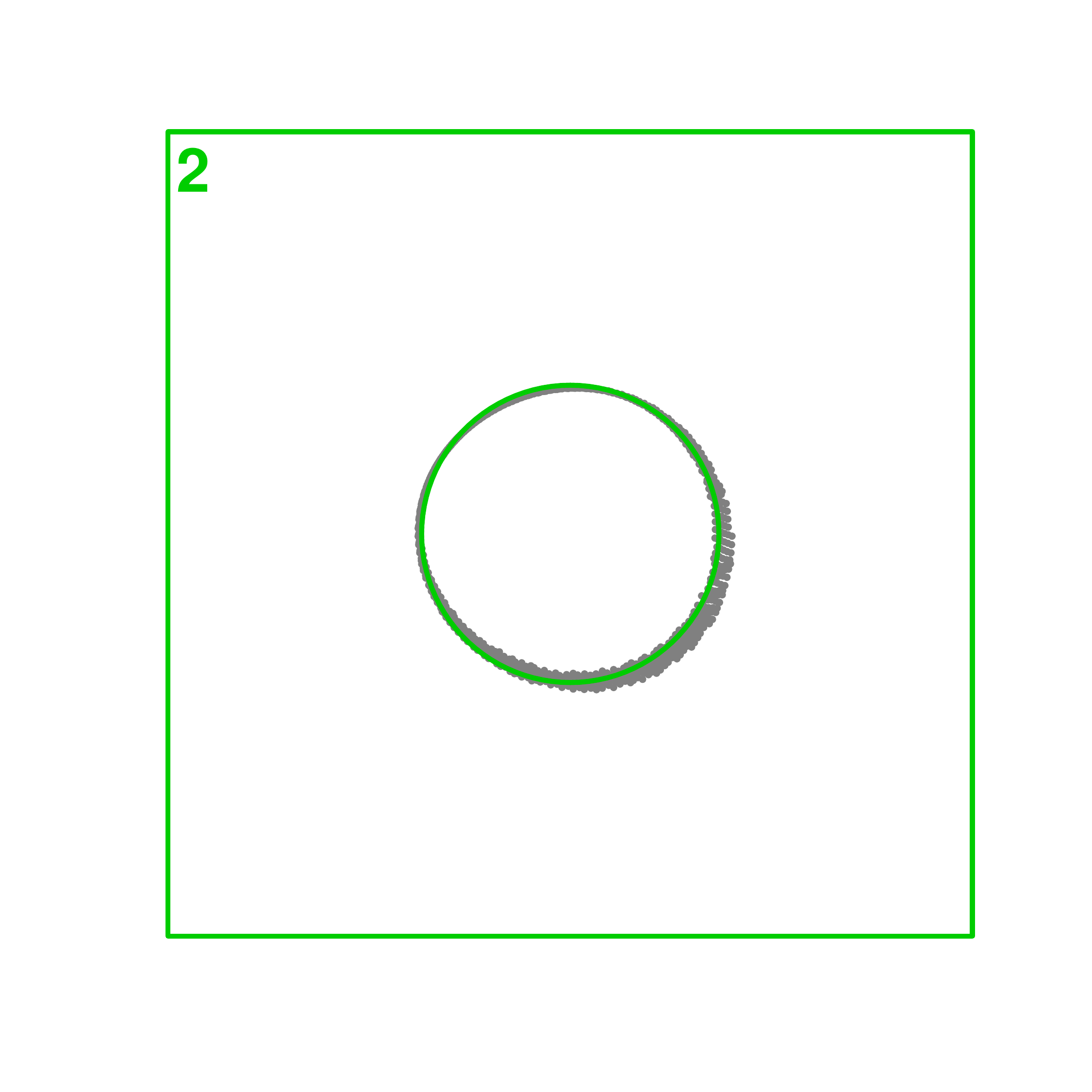}%
            \end{minipage}%
            \begin{minipage}{0.49\textwidth}%
                \includegraphics[width=0.99\textwidth]{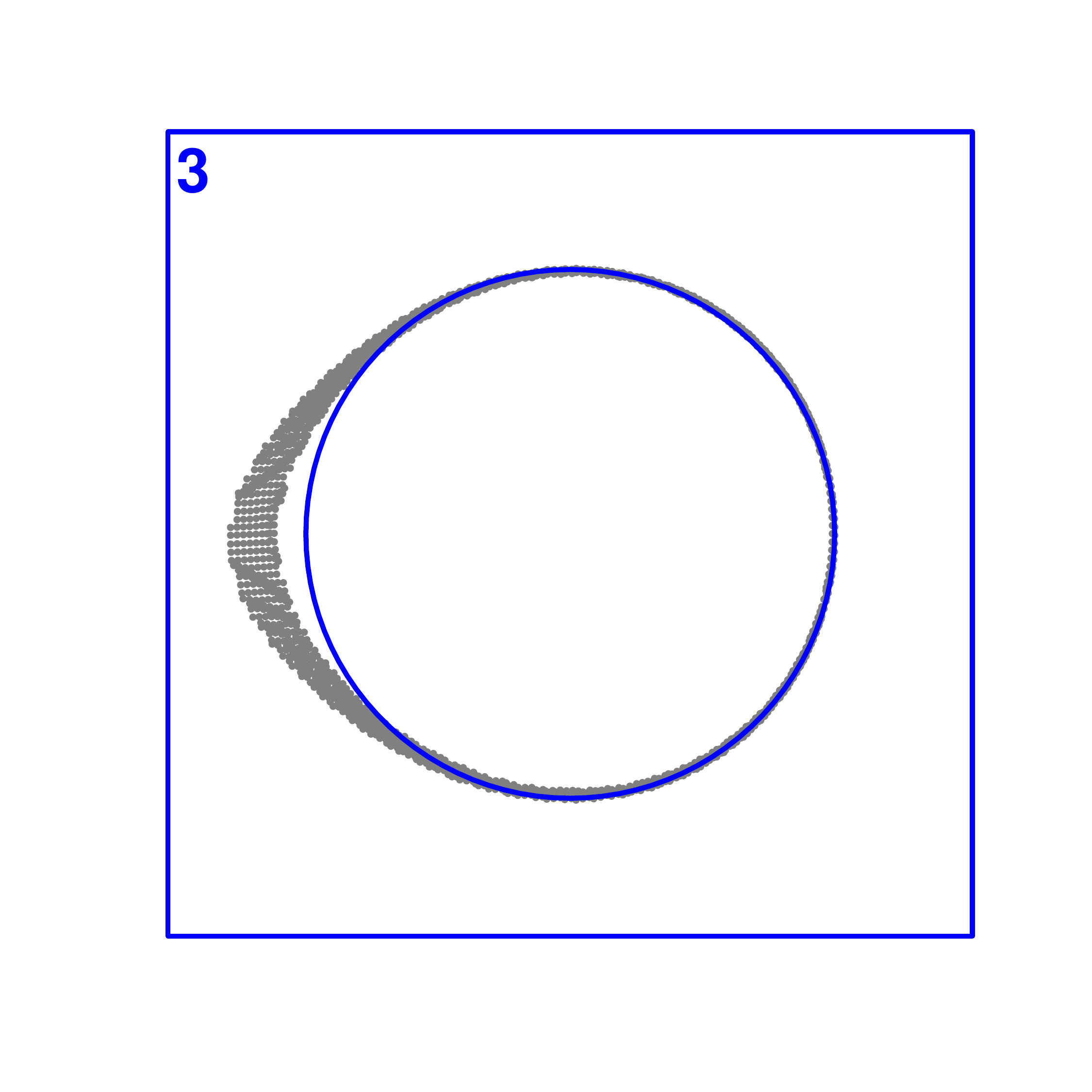}%
                \\%
                \includegraphics[width=0.99\textwidth]{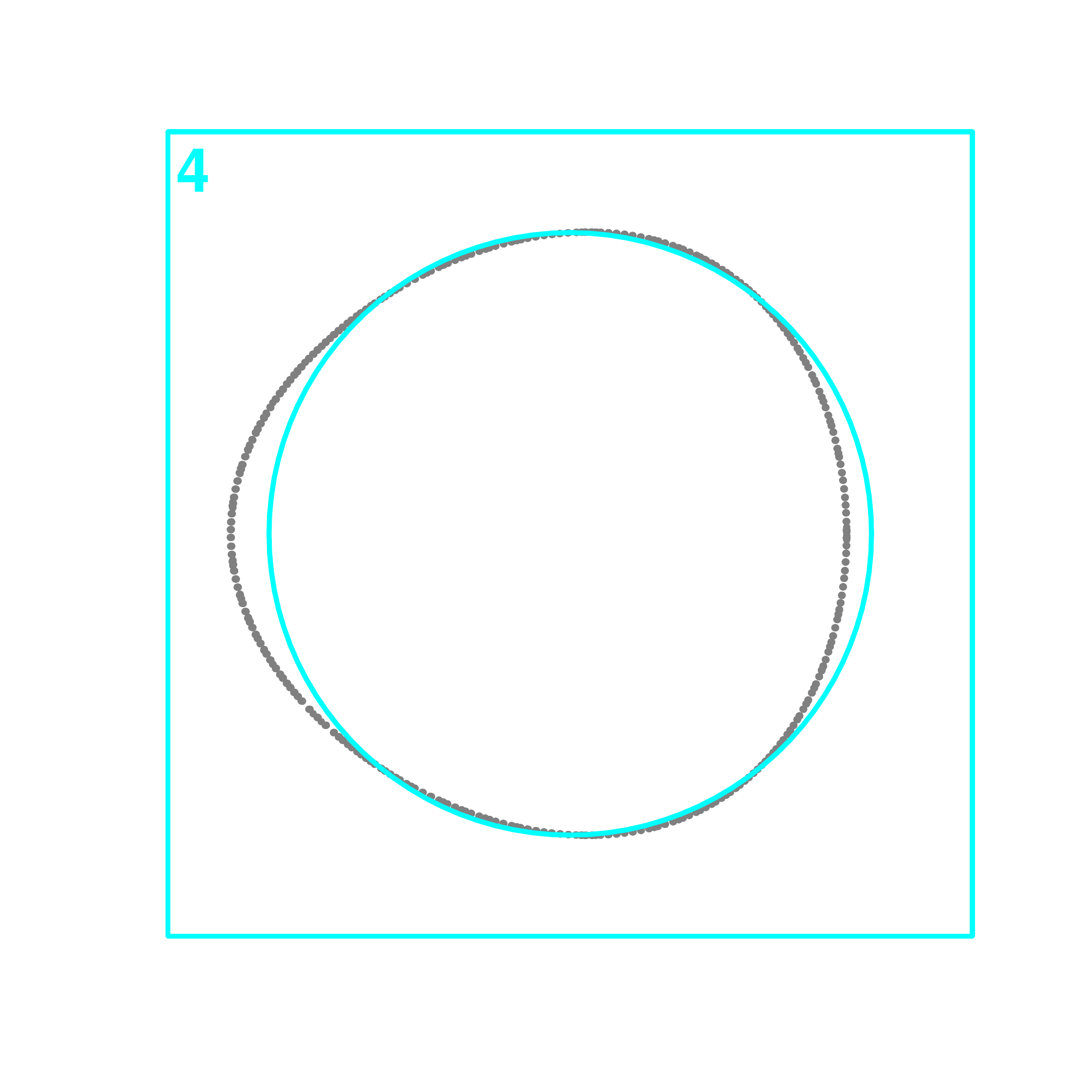}%
            \end{minipage}%
        \end{minipage}%
    }%
    \end{minipage}
    \begin{minipage}{0.99\textwidth}\sidesubfloat[position=bottom]{\includegraphics[width=0.49\textwidth]{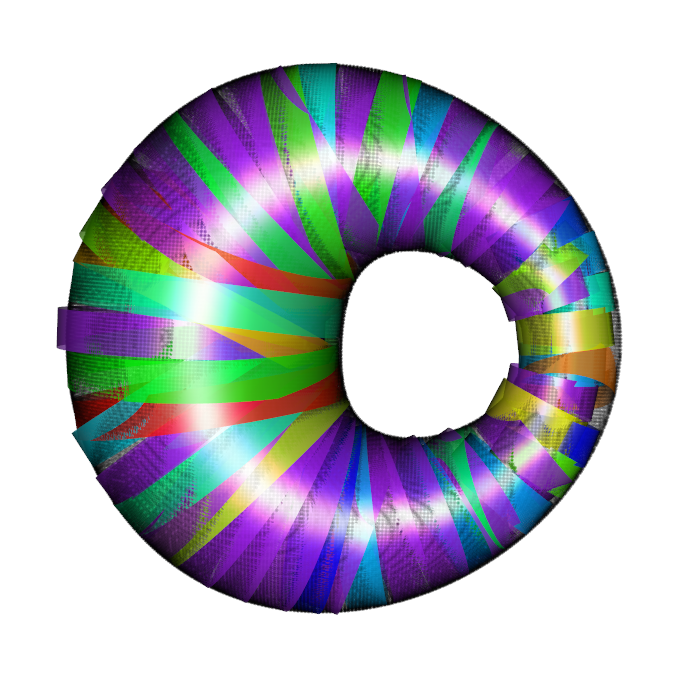} \includegraphics[width=0.49\textwidth]{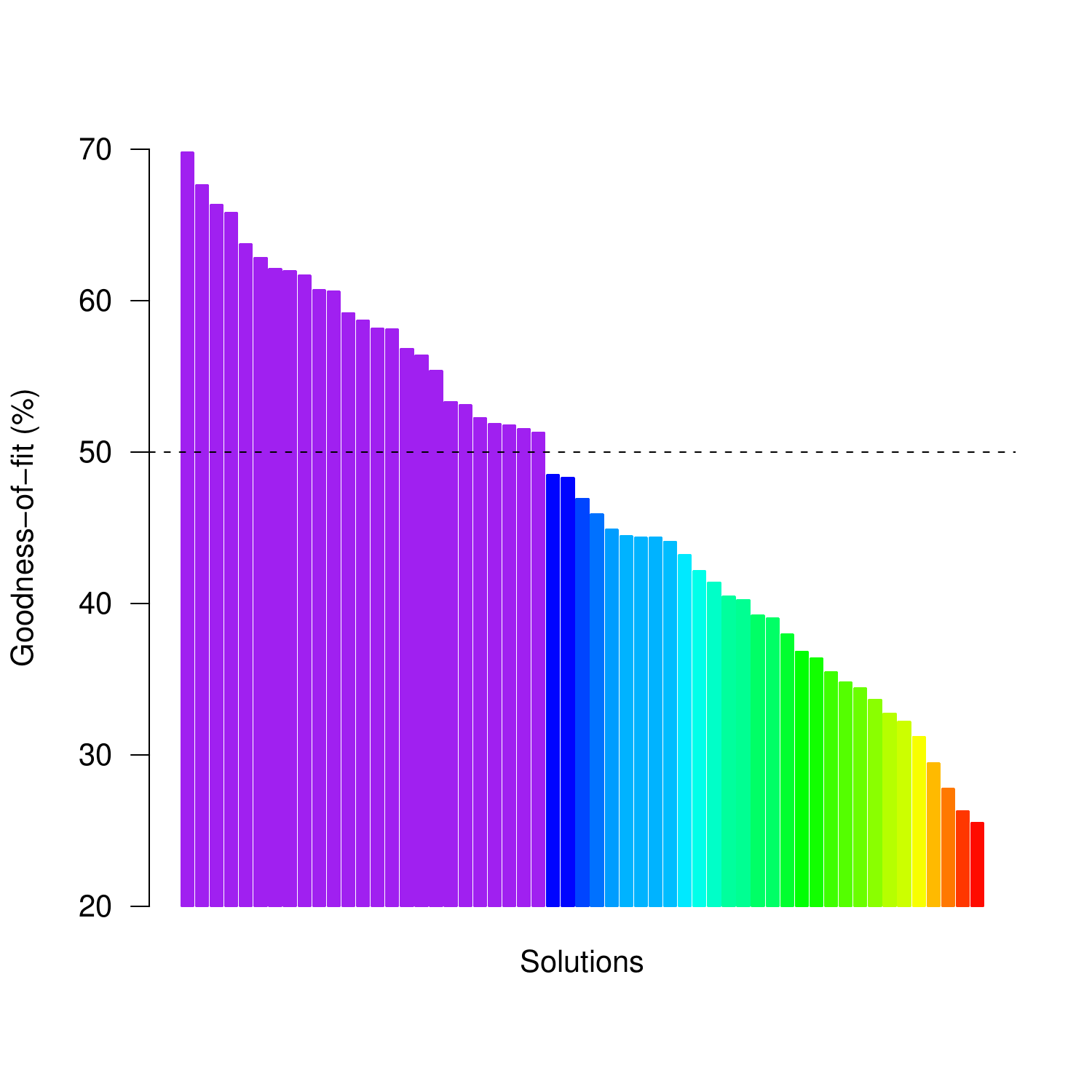}}
    \end{minipage}
    
    \caption{Approximation of a ring cyclide as a collection of cylinders. (A) The ring cyclide displayed at two different viewing angles. (B) Population of cylinders with fitness scores higher than the threshold $\alpha = 50\%$. They span most of the inner ring wall and display radial orientations consistent with the cyclide geometry. The cross-sections of the 4 best-fitting cylinders are shown as insets. (C) Selected, above-threshold solutions in purple (as in panel B) along with sub-optimal solutions from the population whose fitness scores are color-coded from blue to red. When grouped, the solutions cover the whole surface of the cyclide.}
    \label{fig:dupin}
\end{figure*}

We conducted a series of synthetic experiments with point clouds engineered to showcase common problems with 3D reconstructions: noise (Fig. \ref{fig:noise}), occluded components (Fig. \ref{fig:completeness}), and objects whose geometry departs slightly from the primitive shape used (Fig. \ref{fig:dupin}). The first two synthetic experiments involve the fitting of a single cylinder, whereas the third synthetic experiment demonstrates the fitting of numerous cylinders.

The experiment in Fig. \ref{fig:noise} assesses robustness to noise of the algorithm. Noise is a recurring issue in 3D reconstruction \citep{berger2014state} and is manifested at various levels with all photogrammetry software \citep{comparison_paper}. This experiment features a single cylinder onto which a uniform point jitter was applied, with an amplitude reaching up to 40\% of the cylinder's radius. It demonstrates that the algorithm is robust to noise, as it achieves near-perfect convergence even in the high-noise scenario (Fig. \ref{fig:noise}A,B). It also illustrates how the fitness score is linked to the point cloud quality: the perfectly matching fit obtained with 30\% jitter achieves a fitness score of 0.5, whereas the score without noise is 1 (Fig. \ref{fig:noise}C).

Fig. \ref{fig:completeness} exemplifies how the shape recovering capability is impacted by point cloud completeness. Various levels of completeness are simulated by keeping an arc spanning only a fraction of the full circle, from 90\% down to 10\% (Fig. \ref{fig:completeness}A, B). This mimics the partial point clouds obtained when all photographs are taken from one side of the object of interest. This experiment revealed that cylinders with at least 30\% of their surface present in the point cloud can be fully recovered (Fig. \ref{fig:completeness}B), making the developed approach suitable for reconstructions of partially occluded objects. In the synthetic trials, approximations with 20\% and 10\% completeness converged to local maxima (Fig. \ref{fig:completeness}A, B). It must be noted that these are very hard cases: with less than 20\% of the surface represented, the point cloud amounts to little more than a slightly curved surface, resulting in low fitness scores (Fig. \ref{fig:completeness}C). In addition, convergence was slower compared to the noise experiment, revealing that the search space of incomplete cylinders is harder to explore.

The final example shown in Fig. \ref{fig:dupin} addresses cases where the primitive shape varies from from the modeled object. The point cloud was obtained by sampling regularly a ring cyclide (a special case of Dupin cyclide with an ellipto-hyperbolic parameterization, shown in Fig. \ref{fig:dupin}A). This object presents with substantial challenges as the radial cross-sections are not perfectly circular. 
In addition, this experiment requires optimizing a population of cylinders with different radii and orientations. Results of the optimization show successful matching (Fig. \ref{fig:dupin}B). Fig. \ref{fig:dupin}C proves the feasibility of extracting more matching shapes and hints of opportunities to improve the completeness of derived shapes as a post optimization phase.

\subsection{Real case studies}

\begin{figure}
    \centering
    \begin{minipage}{0.99\textwidth}\sidesubfloat[position=bottom]{\includegraphics[width=0.99\textwidth]{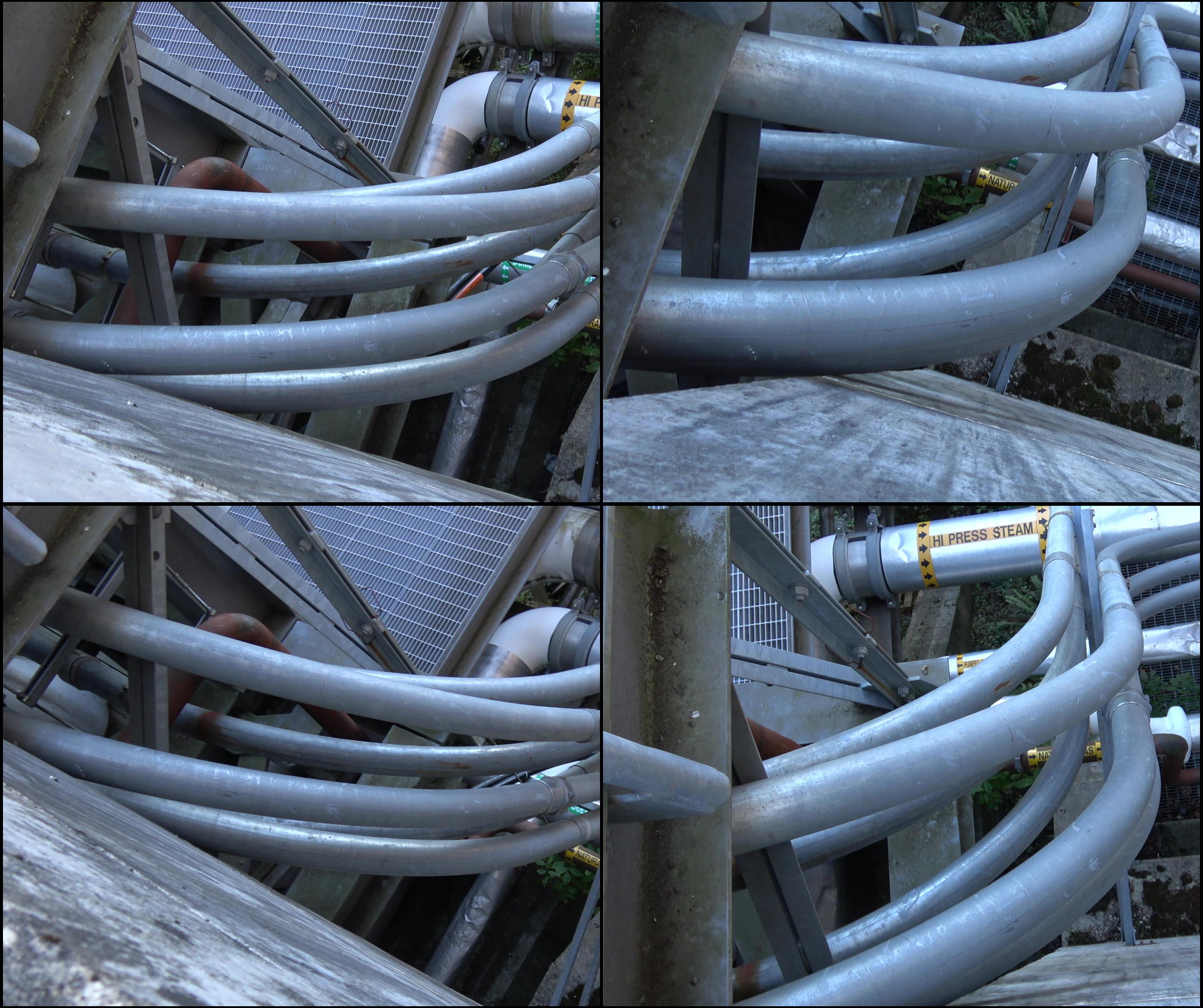}}\end{minipage}
    
    \begin{minipage}{0.94\textwidth}\sidesubfloat[position=bottom]{\includegraphics[width=0.99\textwidth]{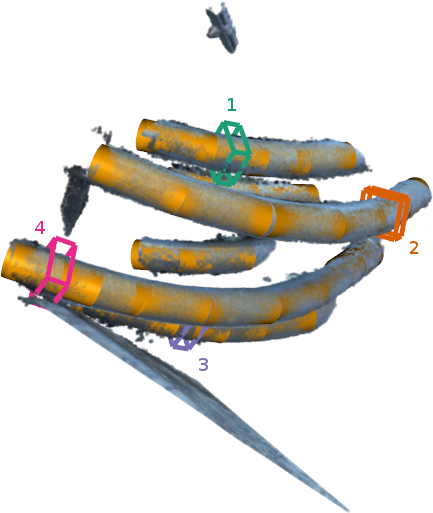}}\end{minipage}
    
    \begin{minipage}{0.99\textwidth}\sidesubfloat[position=bottom]{\begin{minipage}{0.49\textwidth}\includegraphics[width=0.99\textwidth]{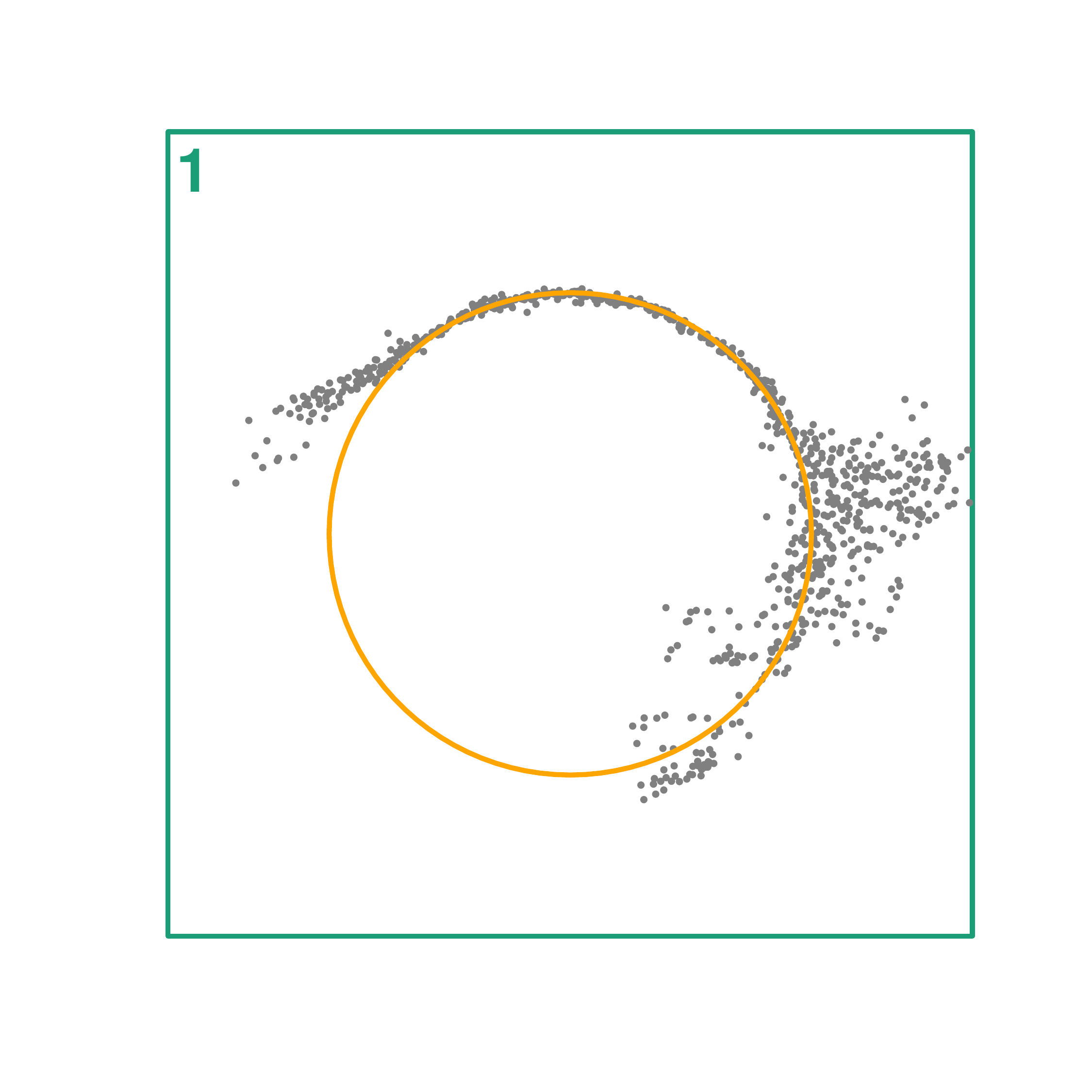}
    
    \includegraphics[width=0.99\textwidth]{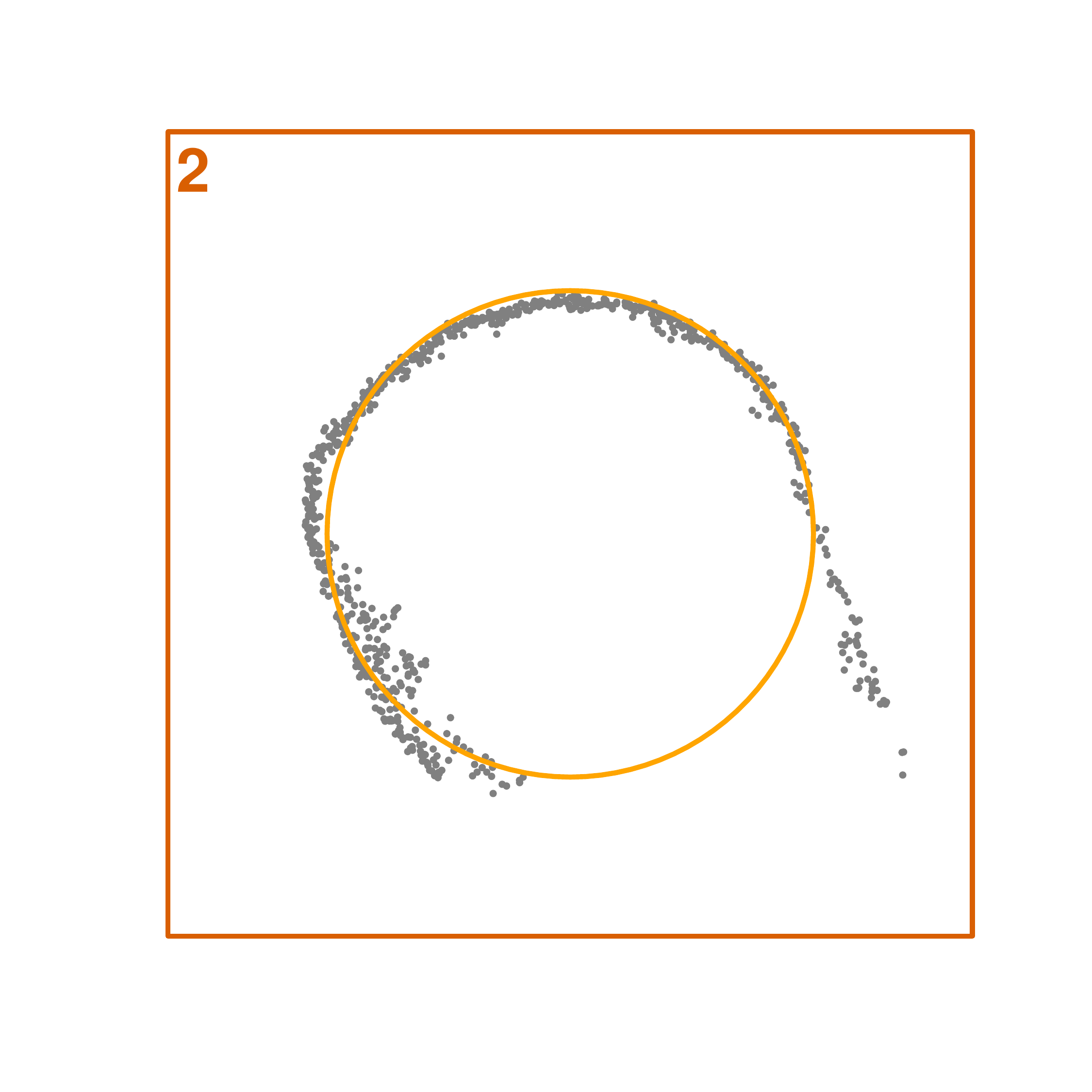}    \end{minipage}\begin{minipage}{0.49\textwidth}\includegraphics[width=0.99\textwidth]{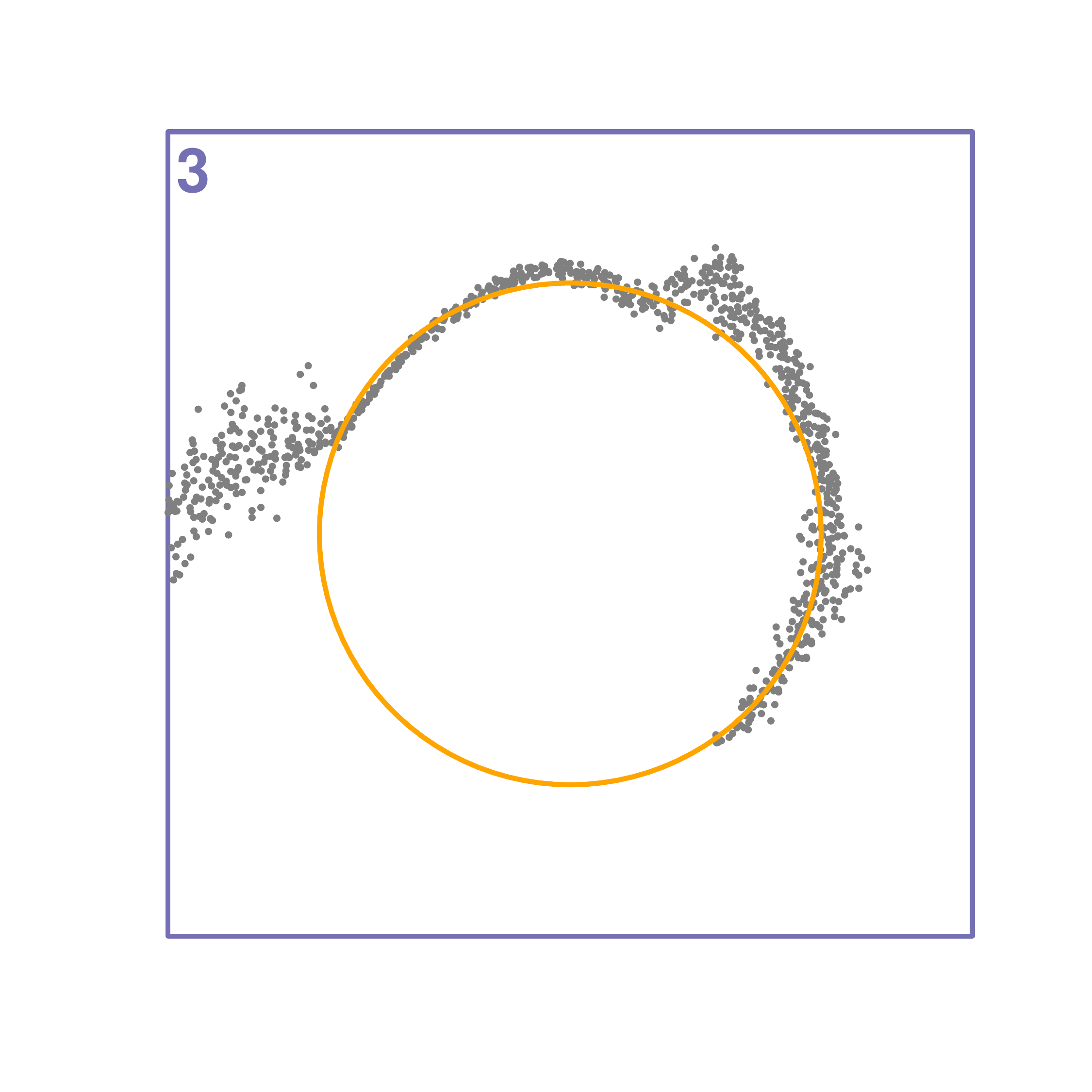}
    
    \includegraphics[width=0.99\textwidth]{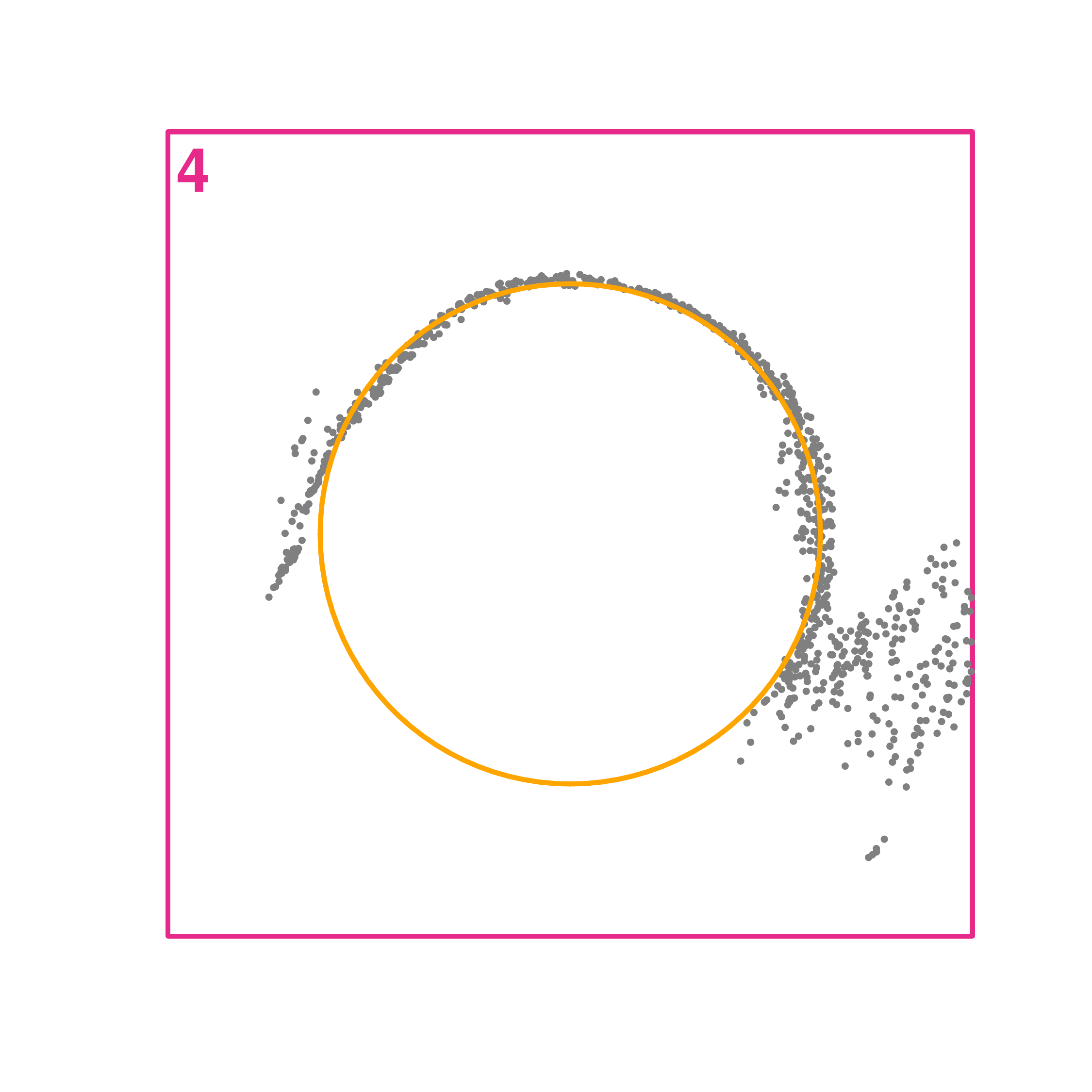}    \end{minipage}}
    \end{minipage}
    \hfill
    \caption{Example of cylinder reconstruction at a pipe-run network. (A) Photographs used for the reconstruction, acquired from an overlooking location and thus limiting the representation to only the upper half of the pipes. 3D point cloud and optimal set of cylinders assigned during the approximation (panel B). 
    (C) Detail of fit cylinders demonstrating adequate recovery of cylindrical shapes even from incomplete point cloud.}
    \label{fig:ohsu}
\end{figure}

\begin{figure*}
    \centering
    \begin{minipage}[b]{0.49\textwidth}\sidesubfloat[position=bottom]{\includegraphics[width=0.9\textwidth]{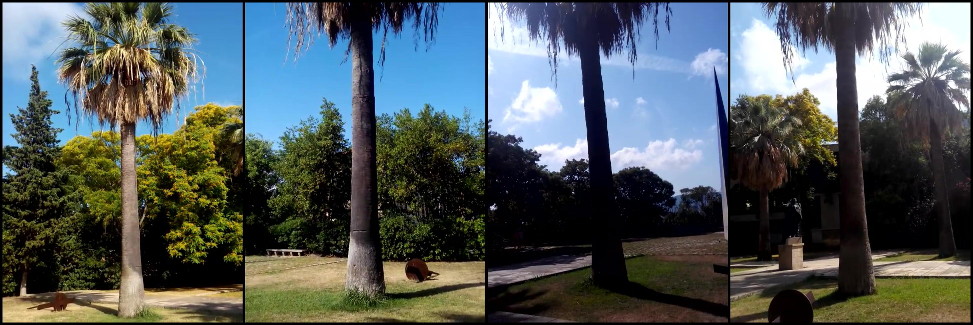}\hfill
    }\end{minipage}
    \begin{minipage}[t]{0.49\textwidth}\sidesubfloat[position=bottom]{\includegraphics[width=0.9\textwidth,trim={0 0 0 3cm},clip]{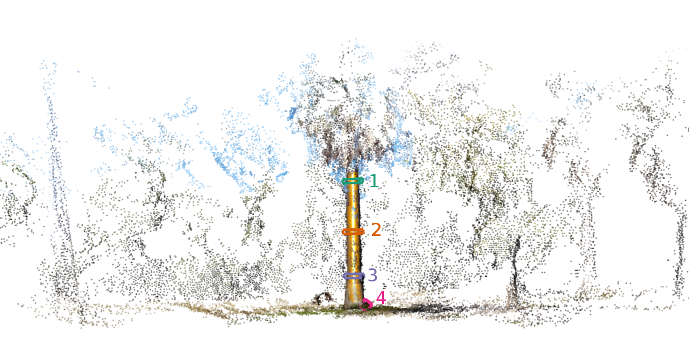}\hfill
    }\end{minipage}

    \begin{minipage}{0.49\textwidth}\sidesubfloat[position=bottom]{\includegraphics[width=0.99\textwidth]{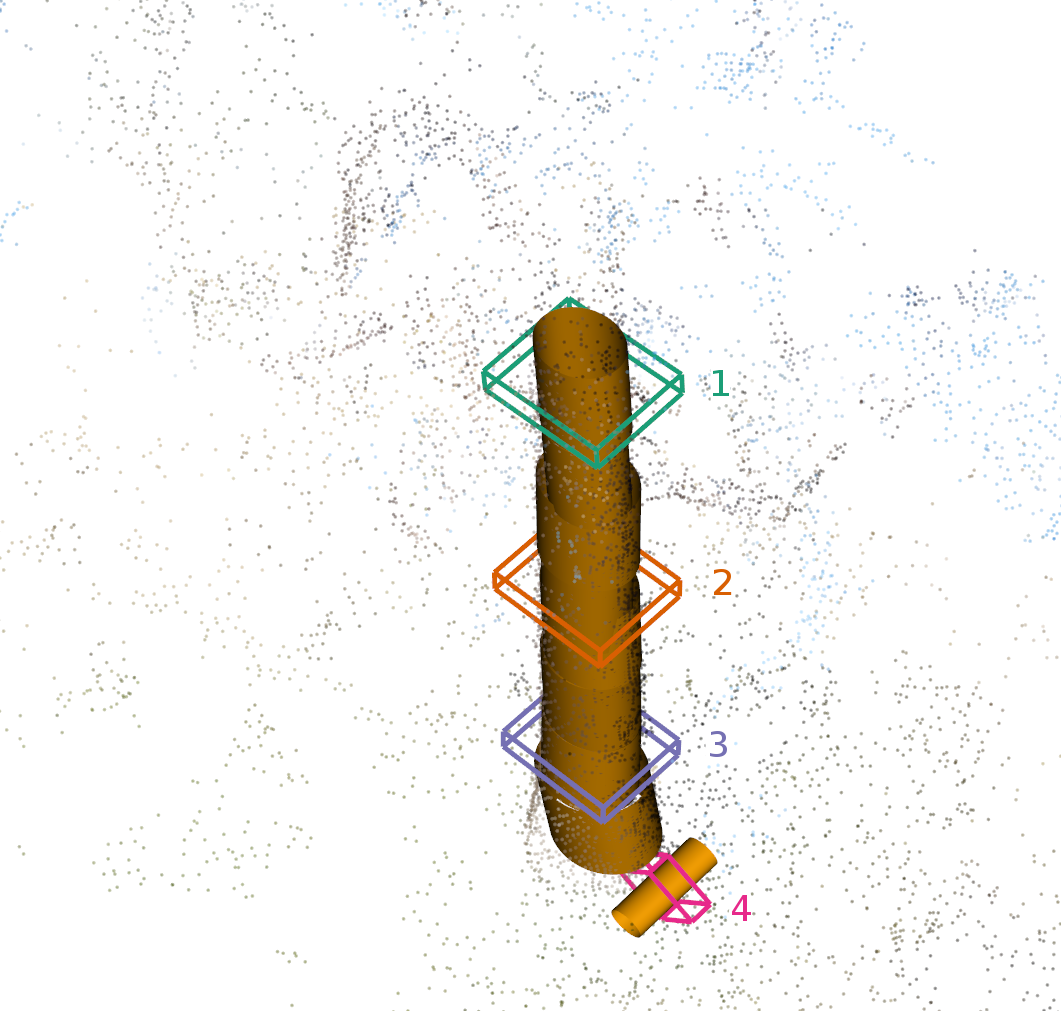}\hfill
    }\end{minipage}
    \begin{minipage}{0.49\textwidth}\sidesubfloat[position=bottom]{\begin{minipage}{0.49\textwidth}\includegraphics[width=0.99\textwidth]{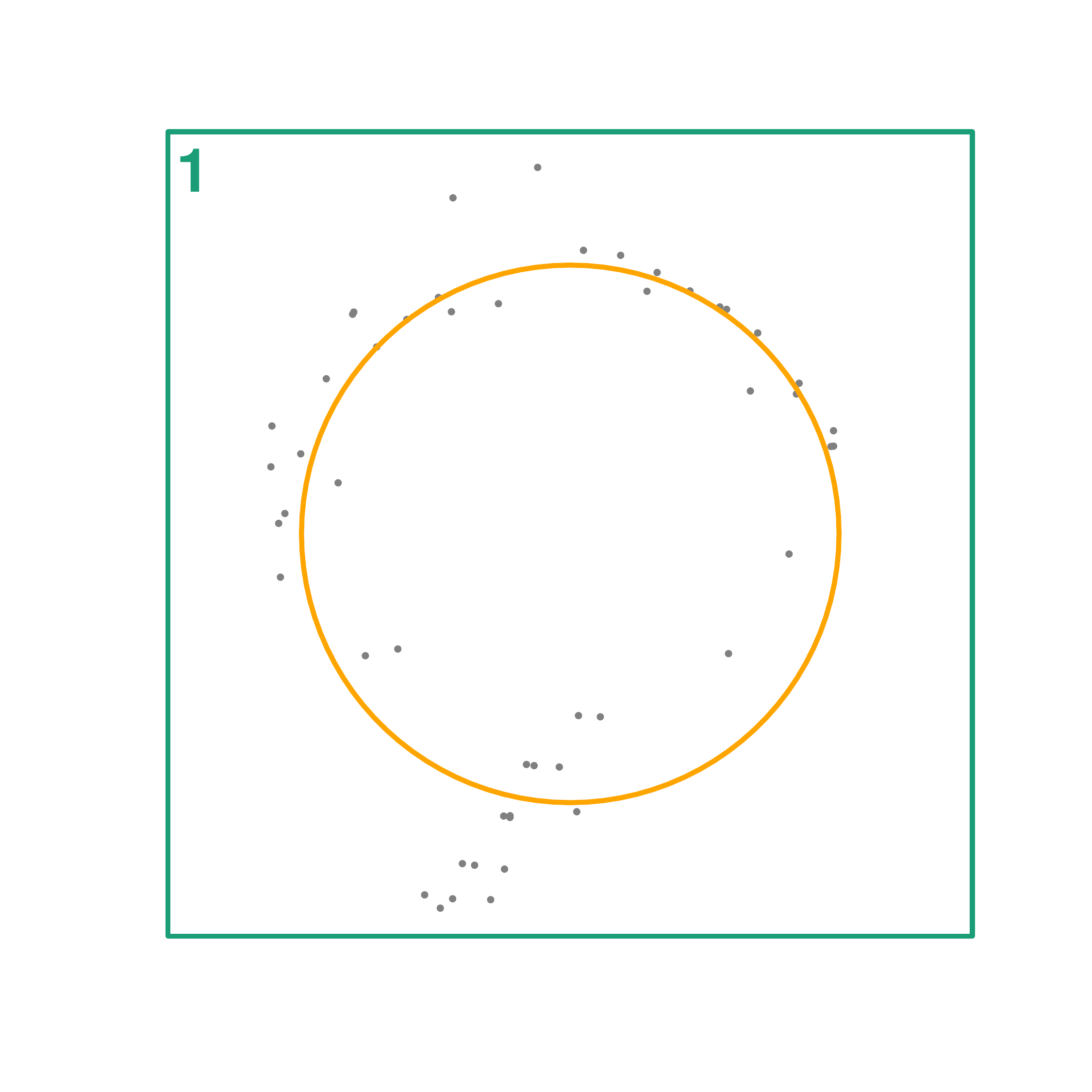}
    
    \includegraphics[width=0.99\textwidth]{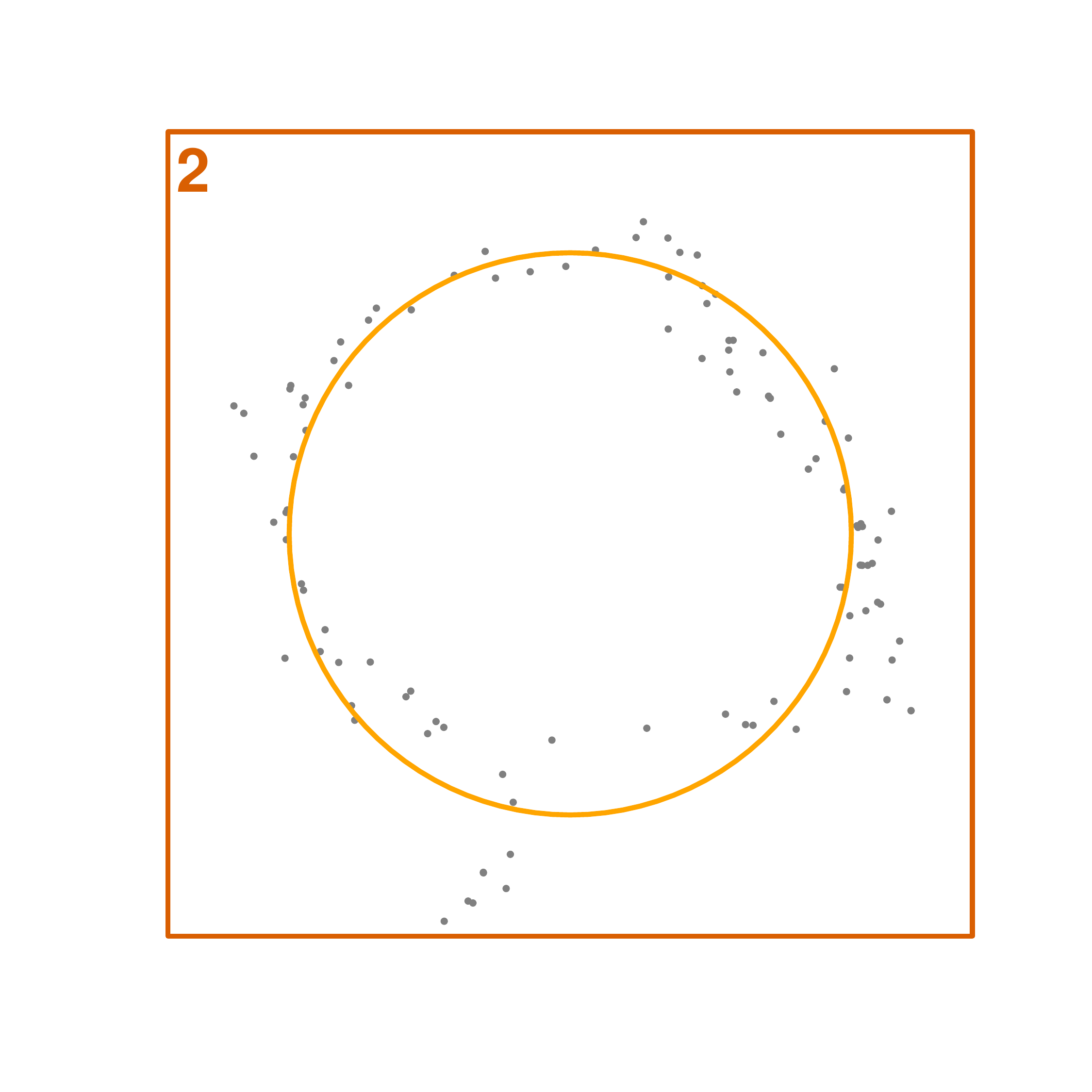}    \end{minipage}\begin{minipage}{0.49\textwidth}\includegraphics[width=0.99\textwidth]{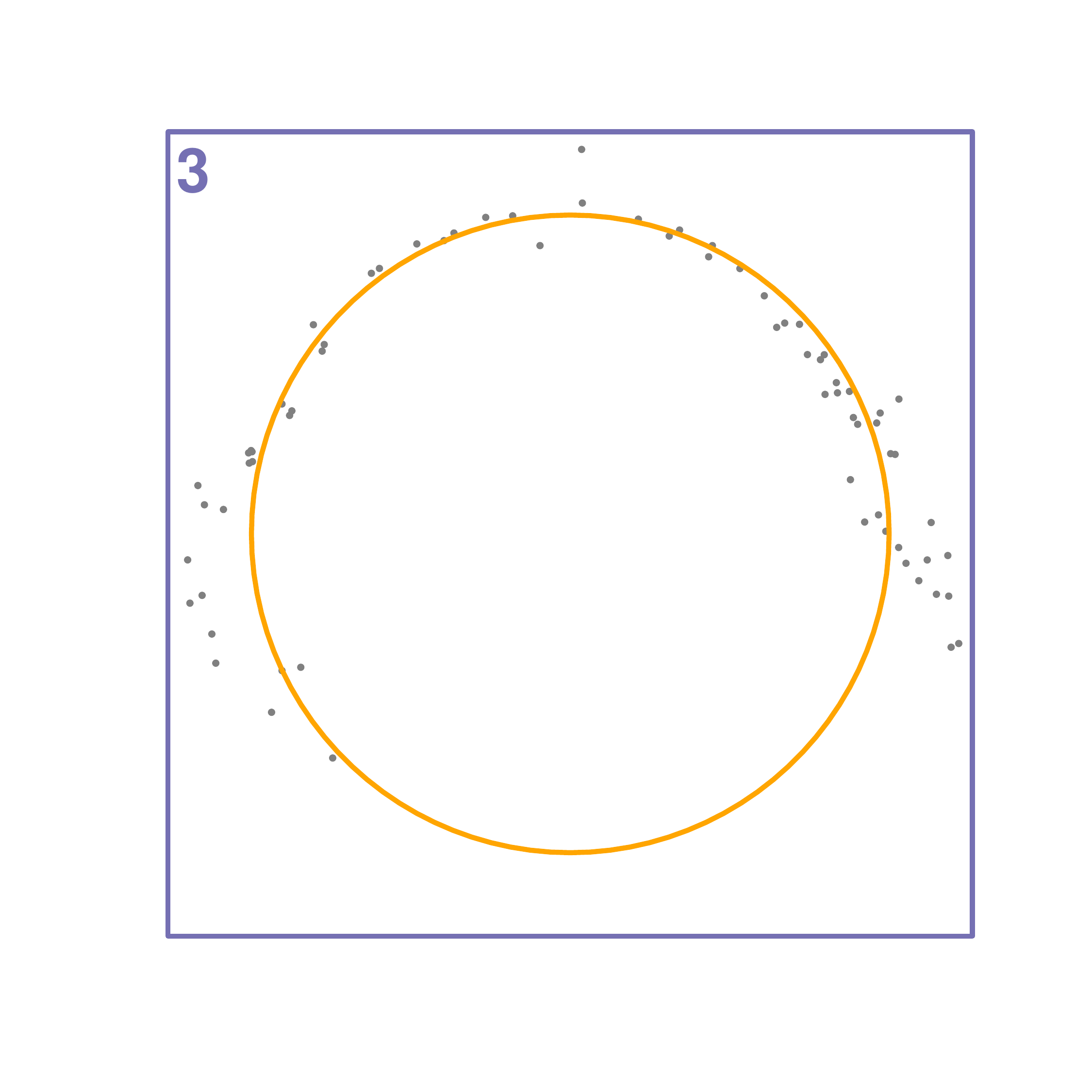}
    
    \includegraphics[width=0.99\textwidth]{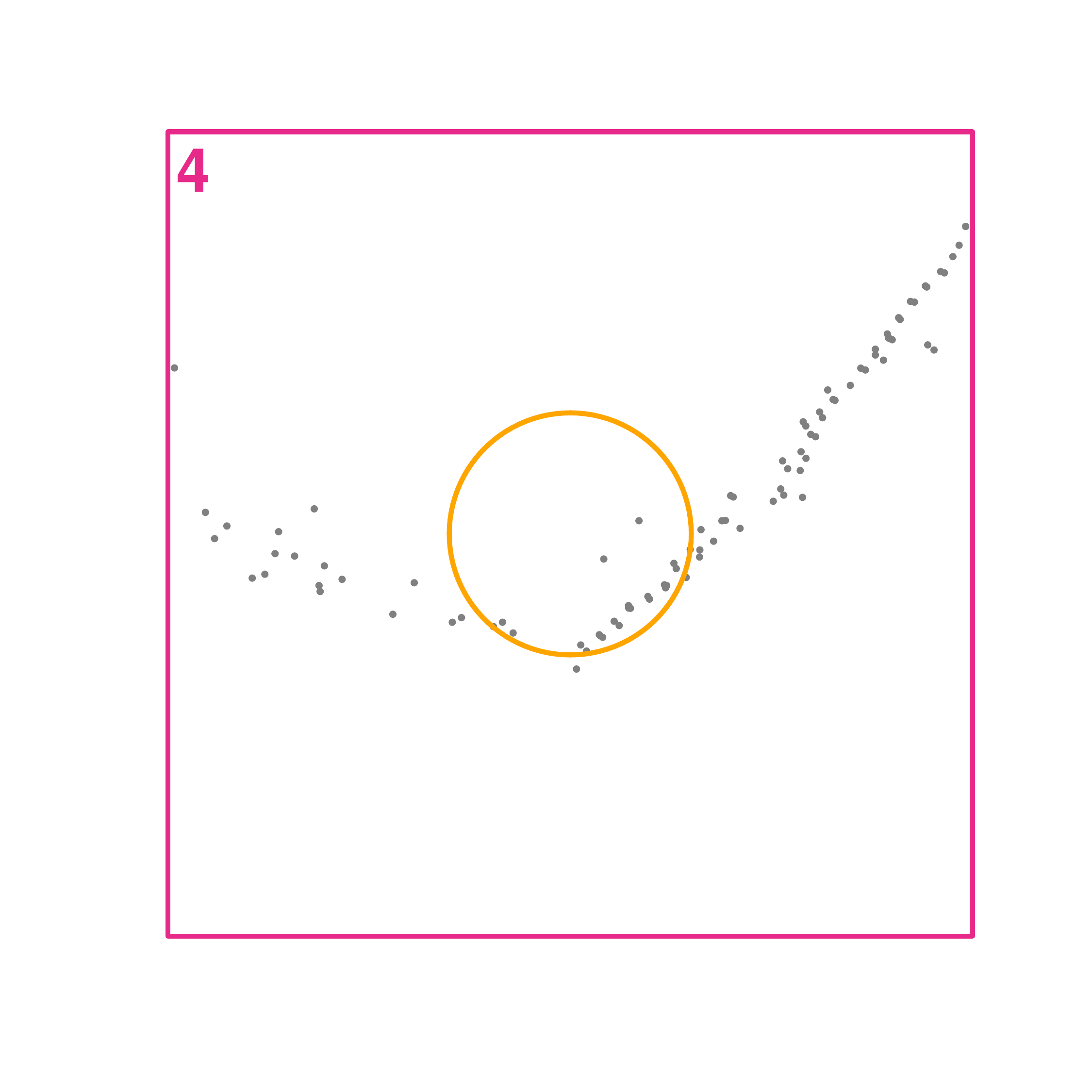}    \end{minipage}}
    \end{minipage}    
    \caption{
    Example of tree stem reconstruction on a test plot with challenging conditions. Photographs acquired along a circle around a targeted tree with a low resolution (640x480 pixels) camera yielded a sparse and noisy point cloud. Panels (A) and (B) show that the entire tree trunk is successfully approximated by a collection of cylinders using a fully automated application of our evolutionary algorithm, without any human intervention or pre-treatment. Cross-sections of the framed area are shown in panel C. Panels C1, C2, and particularly C3 demonstrate that the trunk diameter is captured despite the numerous imperfections of the point cloud. A false positive is found on the ground, close to the tree (captured as frame 4). The cross-section in C4 shows that part of the cylinder does overlap with the points.
    }
    \label{fig:barcelona}
\end{figure*}

We tested our shape optimization algorithm in industrial (Fig. \ref{fig:ohsu}) and vegetation settings (Fig. \ref{fig:barcelona}). The examples were selected to highlight the challenges previously discussed with the synthetic data. In particular, pipes from the industrial pipe-run networks could be imaged only from one side (Fig. \ref{fig:ohsu}A), resulting in a partial reconstruction similar to the second synthetic example (Fig. \ref{fig:completeness}).
Furthermore, cylinders are curved and thus an imperfect match to the chosen primitive shape (as in the Dupin cyclide example of Fig. \ref{fig:dupin}). Despite these challenges, the 3D structure was successfully approximated by the algorithm (\ref{fig:ohsu}B-D). 

Additional challenges arise from the low-density point clouds and the substantial amount of noise visible in the case of tree reconstruction (Fig. \ref{fig:barcelona}). These result from the low, VGA resolution of the camera used. In addition, about one-fourth of the
photos were over-exposed, leading to wide deformations of the corresponding side of the trunk. 
Despite these challenges, the algorithm succeeded in fitting cylinders to the trunk of the central tree (Fig. \ref{fig:barcelona}, panel D1 to D3). Unlike other approaches that require pre-processing of the point cloud to remove planes and other non-cylindrical  objects \citep[e.g.][]{qiu2014pipe}, we did not perform any modification or cleaning of the initial point cloud obtained by photogrammetry. The vast majority of the points were thus not identified by the algorithm as belonging to a cylinder (Fig. \ref{fig:barcelona}B). Only a single false-positive cylinder was obtained (Fig. \ref{fig:barcelona}, panel D4). An important feature of our algorithm is its ability to adjust \textit{a posteriori} the shape acceptance threshold to remove false detections, as is the case here, or to include more cylinders (as in Fig. \ref{fig:dupin}C).

\section{Discussion}

Point clouds obtained from photographs via photogrammetry are imperfect models of reality and they contain noise, deformations, and have completeness issues. The objects represented rarely have regular shapes. Our evolutionary optimization method largely resolves these limitations, is able to approximate and recover the underlying object geometry from point clouds, and can capitalize on \textit{a priori} knowledge of the geometric structure of scene objects. To the best of our knowledge, this is the first time a genetic strategy has been deployed to evolve the geometric properties of shapes. We extended the classical evolutionary paradigm on real-valued encodings with the design of a set of spatial mutation operators, and we analyzed their contribution to the overall optimization performance. We have also identified a number of limitations associated with shape optimization in 3D, which we investigated through a series of synthetic experiments. Finally, we demonstrated application of our approach to a set of actual examples.

Our optimization procedure is based on evolutionary algorithms and tackles the challenging problem of shape approximation from an incomplete 3D scene. 
Compared to general purpose mesh reconstruction procedures (such as Ball-pivoting algorithm or Poisson surface Reconstruction, cf. graphical abstract), our shape-based method supports recovering the 3D structure of even partially-occluded objects.
Because it does not require an iterative segmentation of the scene to isolate potential cylinder locations \citep[such as][]{schnabel2007efficient}, our method avoids artifacts leading to false positives (identification of non-existent cylinders) and false negatives (failure to identify existent cylinders). Unlike other approaches that rely on heuristics requiring complete cylinders \citep{qiu2014pipe}, our method is able to effectively recover cylindrical shapes from partially occluded objects (Fig. \ref{fig:completeness}).

This study introduces a novel evolutionary algorithm able to fit a collection of shapes to a set of 3D points. 
Although we demonstrated the optimization of cylindrical shapes, our method can actually be applied to any shape. In particular, our method could be used to recovers 3D scene as a collection of composite shapes, pursuing essentially a goal similar to the primitive fitting approach of \cite{schnabel2007efficient}. One could expect from this potential extension of our work, the same trade-offs observed with cylinders: namely, the ability to obtain a superior model accuracy at the expense of higher computational power.
3D shapes require rigorous parameterization and depend on a set of customized mutation operators capable of efficiently exploring the search space. We proposed a set of such operators and we demonstrated how to rank them based on concepts originating from the game theory (relative importance derived from Shapley values). In the interest of manipulating several types of primitive shapes at once, it is desirable to design a mutation operator that transforms a primitive shape of one type into the closest geometrical shape of another type (for example, a sphere into a cuboid). Further investigation on this topic is required.

Keeping the search global comes at computational price. Whereas typical shape recovery with RANSAC-based algorithms uses seconds-to-minutes on a standard computer, the genetic approach described here takes minutes-to-hours - up to one day for large, high-density point clouds. 
This is on par with the computational cost required to obtain the point clouds from pictures using photogrammetry software.
In addition, improvements targeting optimization speed-ups are possible. Lowering the point density with an intelligent thinning operation can lead to important performance gains. The iterative initiation of optimization runs on increasingly denser point clouds could further relay optimal parameters already identified in prior runs to improve performance. Such a sequential fitting approach could be improved by adding a transferability objective relying on a surrogate model \citep[here, the less sampled point cloud; see also][]{pinville2011promote,koos2013transferability}. 
Alternatively, coupling the efficient but locally-based RANSAC search with our time-consuming but global genetic search is a  promising idea. A simple hybrid scheme could involve the segmentation of the point cloud using the best-fitting cylinders of the genetic search, and the local search of cylinders with the RANSAC approach.

\section*{Acknowledgments}

I am thankful to Nikolay Strigul and Demetrios Gatziolis for acquiring financial support (joint-venture agreement between the USDA Forest Service Pacific Northwest Research Station and Washington State University) and for proofreading. DG formulated the overarching research goal of using Structure-from-Motion technology to assess understory vegetation dimensions.

\bibliographystyle{apalike} 
\bibliography{ref.bib}

\end{document}